\documentclass[onecolumn, 10pt]{IEEEtran}
\usepackage{amsmath,amssymb,dsfont}
\usepackage{graphicx}      % include this line if your document contains figures
\usepackage{psfrag,graphicx,epsfig}
\usepackage{epstopdf}
\usepackage{xspace}
\usepackage{subfig}
\usepackage{float}
\usepackage{placeins}
\usepackage{multirow}
\usepackage{pgf,tikz}
\usepackage{nowidow}
\usepackage{siunitx}
\usepackage{color}
\usepackage{flushend}
\usepackage[vlined,linesnumbered,boxruled]{algorithm2e}
\usepackage{algpseudocode}

\usepackage{etex}
\usepackage{array}
\usepackage{cite}
\newcolumntype{x}[1]{%
	>{\centering\hspace{0pt}}p{#1}}%

\newcommand{\EQ}{\begin{eqnarray}}
	\newcommand{\EN}{\end{eqnarray}}
\newcommand{\EQQ}{\begin{eqnarray*}}
	\newcommand{\ENN}{\end{eqnarray*}}

\renewcommand{\natural}{\mathds{N}}

\newcommand{\indicator}[2]{\operatorname{I}_{#1}(#2)}

\usepackage{bm}
\newcommand{\trace}[1]{\text{trace}\left(#1\right)}
\newcommand{\card}[1]{\left|#1\right|}
\newcommand{\abs}[1]{\left|#1\right|}
\newcommand{\eigval}{\lambda}

\newcommand{\prob}{\text{Pr}}

\newcommand{\loglike}{\mathcal{L}}

\newcommand{\pfsa}{G} %M
\newcommand{\state}{q}
\newcommand{\stateVar}{Q}
\newcommand{\stateSet}{\mathcal{Q}}
\newcommand{\alphabet}{a}
\newcommand{\alphabetSet}{\mathcal{A}}
\newcommand{\symb}{s}
\newcommand{\symbVar}{S}
\newcommand{\symbSeq}{\vec{\symb}}
\newcommand{\trFn}{\delta}
\newcommand{\emProb}{m} %\tilde{\pi}

\newcommand{\emMat}{\bm{M}} %\bm{\tilde{\Pi}}
\newcommand{\depth}{D}
\newcommand{\trProb}{\pi}

\newcommand{\trMat}{\bm{\Pi}}
\newcommand{\stProb}{p}
\newcommand{\stProbVec}{\bm{\stProb}}
\newtheorem{theorem}{\bf Theorem}[section]

\newtheorem{definition}{\bf Definition}[section]
\newtheorem{remark}{\bf Remark}[section]
\begin{document}

\title {
{\LARGE \bf
	Symbolic Analysis-based Reduced Order \vspace{3pt}\\ Markov Modeling of Time Series Data
	}%$^\bigstar$}
}

\author{Devesh K. Jha$^{a,1}$,  Nurali Virani$^{a,2}$, Jan Reimann$^{b}$, Abhishek Srivastav$^{c}$, Asok Ray$^{a, b}$
	\thanks {$^a$ Devesh K. Jha, N. Virani and Asok Ray are with Mechanical \& Nuclear Engineering Department, Pennsylvania State University, University Park, PA 16802, USA, and are partially supported by the U.S. Air Force Office of Scientific Research under Grant No. FA9550-15-1-0400; {\tt\{dkj5042,nnv105,axr2\}@psu.edu}}
	\thanks{$^b$ Jan Reimann and Asok Ray are with Department of Mathematics, Pennsylvania State University, University Park, PA, 16802, USA; {\tt\{jan.reimann,axr2\}@psu.edu}. Jan Reimann was partially supported by NSF Grant DMS-1201263}
	\thanks{$^c$ Abhishek Srivastav is with AI and Machine Learning Lab, GE Global Research Center, San Ramon, CA, USA; {\tt srivastav@ge.com}}
	\thanks{$^1$ Currently with Mitsubishi Electric Research Laboratories, Cambridge, MA 02139}
	\thanks{$^2$ Currently with AI and Machine Learning Lab, GE Global Research Center, Niskayuna, NY} 
}

\maketitle
\pagestyle{plain}

%\linenumbers           % Insert line numbers
\begin{abstract}
This paper presents a technique for reduced-order Markov modeling for compact representation of time-series data. In this work, symbolic dynamics-based tools have been used to infer an approximate generative Markov model. The time series data are first symbolized by partitioning the continuous measurement space of the signal and then, the discrete sequential data are modeled using symbolic dynamics. In the proposed approach, the size of temporal memory of the symbol sequence is estimated from spectral properties of the resulting stochastic matrix corresponding to a first-order Markov model of the symbol sequence. Then, hierarchical clustering is used to represent the states of the corresponding full-state Markov model to construct a reduced-order (or size) Markov model with a non-deterministic algebraic structure. Subsequently, the parameters of the reduced-order Markov model are identified from the original model by making use of a Bayesian inference rule. The final model is selected using information-theoretic criteria. The proposed concept is elucidated and validated on two different data sets as examples. The first example analyzes a set of pressure data from a swirl-stabilized combustor, where controlled protocols are used to induce flame instabilities. Variations in the complexity of the derived Markov model represent how the system operating condition changes from a stable to an unstable combustion regime. In the second example, the  data set is taken from NASA's data repository for prognostics of bearings on rotating shafts. We show that, even with a very small state-space, the reduced-order models are able to achieve comparable performance and that the proposed approach provides flexibility in the selection of a final model for representation and learning.
\end{abstract}

\section{Motivation and Introduction}

Hidden Markov model (HMM) is a widely used statistical learning tool for modeling uncertain dynamical systems~\cite{B06}, where the associated temporal data are used to infer a Markov chain with unobserved states. In this setting, the learning task is to infer the states and the corresponding parameters of the Markov chain. In addition to HMM, several other nonlinear techniques have been proposed for Markov modeling of time-series data. Symbolic time-series analysis-based Markov modeling is a recently proposed technique~\cite{R04} where the states of a Markov chain are represented as a collection of words (i.e., symbol blocks, also referred to as memory words) of different lengths, which can be identified from the time-series data on a discrete space with finite cardinality~\cite{SS04, R04, MR14, CL13}. The symbols are created from the continuously varying time-series data by projecting the data to a set with finite cardinality. A common ground among all these tools of Markov modeling as discrete sequences, is that the Markov chain is induced by probabilistic representation of a deterministic finite state auotmaton (DFSA), often called probabilistic finite state automata (PFSA)~\cite{VTDCC05}. While the PFSA-based inference provides a consistent, deterministic graph structure for learning, the deterministic algebraic structure is generally not a very compact representation and may often lead to large number of states in the induced Markov model. To circumvent this problem attempts have been made to reduce the state-space by merging statistically similar states of the model~\cite{MR14}. The problem is, however, that as these models are constructed by partitioning of phase space of the dynamical system, merging states that are statistically similar leads to algebraic inconsistency. On the other hand, if the states are merged to preserve the algebraic consistency, it leads to statistical impurity in the final models (i.e., states which have different statistics could be merged together). Other approaches for state aggregation in Markov chains could be found in~\cite{GPKK15, V12, XSB14}. However, these papers do not consider inference of the Markov model from the data which may not be suitable for analysis of data-driven systems~\cite{D05}. 

The state space for Markov models, created by using symbolic analysis, increases exponentially with increase in memory or order of the symbolic sequence.  Estimating the right memory is critical for temporal modeling of patterns observed in the sequential data. However, some of the states may be statistically similar and thus merging them can reduce the size of state-space. This paper presents reduced-order Markov modeling of time-series data to capture temporal patterns, where we estimate the size of temporal memory of the symbolic data using the spectral properties of a PFSA whose states are words of length one~\cite{Srivastav2014, JSMR15}. The constraint of deterministic algebraic structure is not imposed by the end objective, but due to the choice of the data representation model. Thus we propose to merge the states and remove the constraint of deterministic algebraic properties associated with PFSA, where the states of the Markov chain are now collection of words from its alphabet of length estimated in the last step. This state aggregation induces a non-determinism in the finite state model. The parameters of the reduced-order Markov model are estimated by a Bayesian inference technique from the parameters associated with the higher-order Markov model. The final model for data representation is selected using information-theoretic criteria, and thus, we get a unique stopping point to terminate the state-merging procedure. We also present a bound on the distortion of the predictive capability of the models up on reduction in the size of the state-space. The final model obtained is a generative model for the data; however, some predictive capability is lost as we remove the deterministic algebraic structure of a DFSA.

The proposed technique of state merging is inspired by time-critical applications where it is imperative to arrive at a reliable decision quickly as the dynamics of the process being monitored is really fast. In such applications, there are strict constraints on accuracy as well as the time needed to come to a decision. In this paper, we illustrate the concepts using two different datasets. We discuss in detail the example of combustion instability which is a highly nonlinear and complex phenomena and results in severe structural degradation in jet turbine engines. Some good surveys on the current understanding of the mechanisms for the combustion instability phenomena could be found in~\cite{OAL15, SSDC03, CDSBM14, HY09, MBDSC12}. Active combustion instability control (ACIC) with fuel modulation has proven to be an effective approach for reducing pressure oscillations in combustors~\cite{BMJK06, BMH07}. Based on the work available in literature, one can conclude that the performance of ACIC is primarily limited by the large delay in the feedback loop and the limited actuator bandwidth ~\cite{BMJK06, BMH07}. Early detection of combustion instability can potentially alleviate the problems with delay in the ACIC feedback loop and thus possibly improve the performance. Some recent work for detection and prediction of combustion instabilities could be found in~\cite{JSR16, VJR16, SCRR16, NTS14, MS15}. While the results in these papers are encouraging, there is no interpretation of the expected changes in the data-driven model that could be observed during changes in the operating regime of the underlying process. In contrast to the work reported in literature, we have presented an overall idea of changes in the underlying stochastic model structure and parameters during the complex instability phenomenon.

\textbf{Contributions.} This paper presents a technique for Markov modeling of time series data using a PFSA with nondeterministic algebraic structure. Nondeterminism is induced by merging states of a PFSA with deterministic algebraic structure inferred from discrete sequential data, which in turn allows very compact representation of temporal data. In contrast to the approach in~\cite{MR14}, we present a method to use information-theoretic criteria to arrive at a consistent stopping criterion for model selection. The resulting reduced-order model has fewer parameters to estimate; this is turn leads to faster convergence rates and thus faster decisions during test (or operation). We also present a bound on the distortion in the predictive capability of the models due to state-space reduction using Hamming distance between the sequences generated by the original and final model. The algorithms presented in the paper are validated on two different datasets-- pressure data obtained from a swirl-stabilized combustor to monitor thermo-acoustic instability and a public data set for bearing prognostics. We show changes in the complexity of the pressure data as the process moves from stable to unstable through the transient phase which is then used to arrive at a criterion that provides perfect class separability. Apart from the results on Markov modeling, the results on combustion instability could be of independent interest in combustion community.

\section{Background and Mathematical Preliminaries}

\begin{figure} %Fig 01
	\centering
	\includegraphics[width=0.9\textwidth]{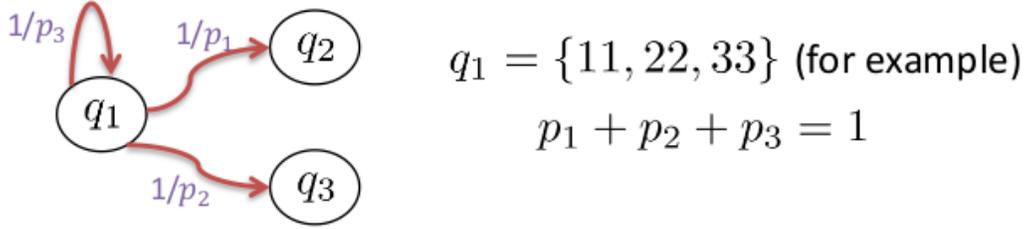}
	\caption{Graphical model showing non-determinism in a PFSA. The symbol $1$ emitted from state $q_1$ leads to different states with fixed probabilities indicating non-deterministic behavior.}
	\label{fig:NonDeterminism}
\end{figure}

Symbolic analysis of time-series data is a recent approach where continuous
sensor data are converted to symbol sequences via partitioning of the continuous
domain~\cite{SAX07, R04}. The dynamics of the symbols sequences are
then modeled as a probabilistic finite state automaton (PFSA), which is defined
as follows:

\begin{definition}[PFSA]\label{defn:PFSA}
	A probabilistic finite state automaton (PFSA) is a tuple $\pfsa =( \stateSet, \alphabetSet, \trFn, \emMat)$ where
	\begin{itemize}
		\item $\stateSet$ is a finite set of states of the automata;
		\item $\alphabetSet$ is a finite alphabet set of symbols $\alphabet \in \alphabetSet$;
		\item $\trFn: \stateSet \times \alphabetSet \rightarrow \stateSet$ is the state transition function;
		\item {$\emMat: \stateSet \times \alphabetSet \rightarrow [0, 1]$ is the
			$\card{\stateSet}\times\card{\alphabetSet}$ emission matrix. The matrix
			$\emMat = [\emProb_{ij}]$ is row stochastic such that $\emProb_{ij}$ is the
			probability of generating symbol $\alphabet_{j}$ from state $\state_{i}$}.
	\end{itemize}
\end{definition}
\begin{remark} The PFSA defined above has a deterministic algebraic structure which is governed by the transition function $\trFn$; thus a symbol emission from a particular state will lead to a fixed state. However, the symbol emissions are probabilistic (represented by the emission matrix). On the other hand, the transition function for a non-deterministic finite state automaton is given by a map, $\trFn: \stateSet \times \alphabetSet \rightarrow 2^\stateSet$ where, $2^\stateSet$ denotes the power set of $\stateSet$ and includes all subsets of $\stateSet$. The idea is also presented in Figure~\ref{fig:NonDeterminism} where we show that the same symbol can lead to multiple states, however in a probabilistic fashion. This allows more flexibility in modeling at the expense of some predictive accuracy.
\end{remark}

For symbolic analysis of time-series data, a class of PFSAs called the
$\depth$-Markov machine have been proposed~\cite{R04} as a sub-optimal but
computationally efficient approach to encode the dynamics of symbol sequences as
a finite state machine.
\begin{definition} (\textbf{$D$-Markov Machine}~\cite{R04, MR14}) \label{def:D-Markov} A $D$-Markov machine is a statistically stationary stochastic process $S= \cdots a_{-1}  a_{0} a_{1} \cdots $ (modeled by a PFSA in which each state is represented by a finite history of $D$ symbols), where the probability of occurrence of a new symbol depends only on the last $D$ symbols, i.e.,
	\begin{equation}\label{eq:D-Markov}
		\prob(s_n \mid \cdots  s_{n-\depth} \cdots s_{n-1} ) = \prob(s_n \mid s_{n-\depth} \cdots s_{n-1}) \nonumber
		%P[s_n \mid s_{n-1} \cdots s_{n-D} \cdots] = P[s_n \mid a_{n-1} \cdots a_{n-D}]
	\end{equation}
	where $\depth$ is called the depth of the Markov machine.

\end{definition}
A $D$-Markov machine is thus a $\depth^{th}$-order Markov approximation of the discrete symbolic process. For most stable and controlled engineering systems that tend to forget
their initial conditions, a finite length memory assumption is reasonable. The $\depth$-Markov machine is represented as a PFSA and states of this PFSA are words over alphabet $\alphabetSet$ of length $\depth$ (or less);
the state transitions are described by a sliding block code of memory $\depth$
and anticipation length of one~\cite{LM95}.

For systems with fading memory it is expected that the predictive influence of a
symbol progressively diminishes. In this context, depth is defined as follows.
\begin{definition}[Depth]\label{def:depth}
	Let $\symbSeq = \symb_{1}\dots\symb_{k}\symb_{k+1}\symb_{k+2}\dots$ be the
	observed symbol sequence  where each $\symb_{j}\in\alphabetSet\;\forall\; j \in
	\natural$. Then, the depth of the process generating $\symbSeq$ is defined as the
	length $\depth$ such that:
	\begin{equation}\label{eq:depthTrueDefn}
		\prob(\symb_{k}|\symb_{k - 1},\dots,\symb_{1}) = \prob(\symb_{k}|\symb_{k - 1},\dots,\symb_{k - \depth})
	\end{equation}
	
\end{definition}
An accurate estimation of depth for the symbolic dynamical process is required for the precise modeling of the underlying dynamics of the discrete sequence. Next we introduce an information-theoretic metric which is used for merging the states of the Markov model later in next section.
\begin{definition}[Kullback-Leibler Divergence]\label{def:KLD}~\cite{G90}
	The Kullback-Leibler (K-L) divergence of a discrete probability distribution $P$ from another distribution $\tilde{P}$ is defined as follows.
	\begin{equation}
		D_{\textrm {KL}}(P\|\tilde{P})=\sum_{x\in X} {p}_X(x)\log\bigg(\frac{{p}_X(x)}{\tilde{p}_X(x)}\bigg) \nonumber
	\end{equation}
	It is noted that K-L divergence is not a proper distance as it is not symmetric. However, to treat it as a distance it is generally converted into symmetric divergence as follows, $d(P,\tilde{P})= D_{\textrm {KL}}(P\|\tilde{P})+D_{\textrm {KL}}(\tilde{P}\|P)$. This is defined as the K-L distance between the distributions $P$ and $\tilde{P}$.
\end{definition}

This distance is used to find out the structure in the set of the states of the PFSA-based Markov model whose states are words, over the alphabet of the PFSA, of length equal to the depth estimated for the discretized sequence.

\section{Technical Approach}
In this section, we present the details of the proposed approach for inferring a Markov model from the time series data. As discussed earlier, the first step is the discretization of the time-series data to generate a discrete symbol sequence. While it is possible to optimize the symbolization of time-series using some optimization criterion, we do not discuss such a technique here. The data is discretized using the unbiased principle of entropy maximization of the discrete sequence using maximum entropy partitioning (MEP)~\cite{RR06}. The proposed approach for Markov modeling then consists of the following four critical steps
\begin{itemize}
	\item Estimate the approximate size of temporal memory (or order) of the symbol sequence.
	\item Cluster the states of the high-order Markov model.
	\item Estimate the parameters of the reduced-order Markov model (i.e., the transition matrix).
	\item Select the final model using information theoretic scores (described below, Section~\ref{subsec:MDL}).
\end{itemize}
Memory of the discrete sequence is estimated using a recently introduced method based on the spectral analysis of the Markov model with depth $1$. induced by a PFSA~\cite{Srivastav2014, JSMR15}. It is noted that these steps are followed during training to estimate the approximate model for data and during test, the parameters are estimated for the reduced-order model.
% After these three key steps, the final model is selected for representation using a two-stage MDL formulation.
The key ideas behind these steps are explained in the next section.

\subsection{Estimation of Reduced-Order Markov Model}\label{subsec:reducedorder}
Depth $\depth$ of a symbol sequence has been redefined in~\cite{Srivastav2014}
as the number of time steps after which probability of current symbol is
independent of any past symbol i.e.:
\begin{equation}\label{eq:depthDefn}
	\prob(\symb_{k}|\symb_{k - n}) = \prob(\symb_{k})  \ \forall n>\depth
\end{equation}
Note that dependence in the proposed definition (eq.~\ref{eq:depthDefn}) is
evaluated on individual past symbols using $\prob(\symb_{k}|\symb_{k - n})$ as
opposed to the assessing dependence on words of length $\depth$ using  $
\prob(\symb_{k}|\symb_{k - 1},\dots,\symb_{k - \depth})$. It is shown that if
the observed process is {\it forward causal} then observing any additional
intermediate symbols  $\symb_{k- 1},\dots,\symb_{k -  n + 1}$ cannot induce a
dependence between $\symb_{k }$ and $\symb_{k - n}$ if it did not exist on
individual level~\cite{Srivastav2014}.

Let $\trMat = [\trProb^{(1)}_{ij}]$ be the one-step transition probability
matrix of the PFSA $\pfsa$ constructed from this symbol sequence i.e.
\begin{equation}\label{eq:stateTransition1step}
	\trMat = \prob(\symb_{k}|\symb_{k - 1})
\end{equation}
Then using the distance of the transition matrix after steps from the stationary
point, depth can be defined as a length $\depth$ such that
\begin{equation}\label{eq:depthTrace}
	\abs{\trace{\trMat^{n}} - \trace{\trMat^{\infty}}} \leq \sum_{j = 2}^{J}
	\abs{\eigval_{j}}^n < \epsilon \ \forall n > \depth
\end{equation}
where $J$ is number of non-zero eigenvalues of $\trMat$. Thus, the depth $\depth$ of the symbol sequence is estimated for a choice of $\epsilon$ by estimating the stochastic matrix for the one-step PFSA. Next, another pass of data is done to estimate the PFSA parameters whose states are words over $\alphabetSet$ of length $\depth$, i.e., $\trMat = \prob(\symb_{k}|\symb_{k - 1},\dots, \symb_{k - D})$. It is noted that this step is critical for modeling accuracy.

\begin{figure} %Fig 02
	\centering
	\includegraphics[width=0.75\textwidth]{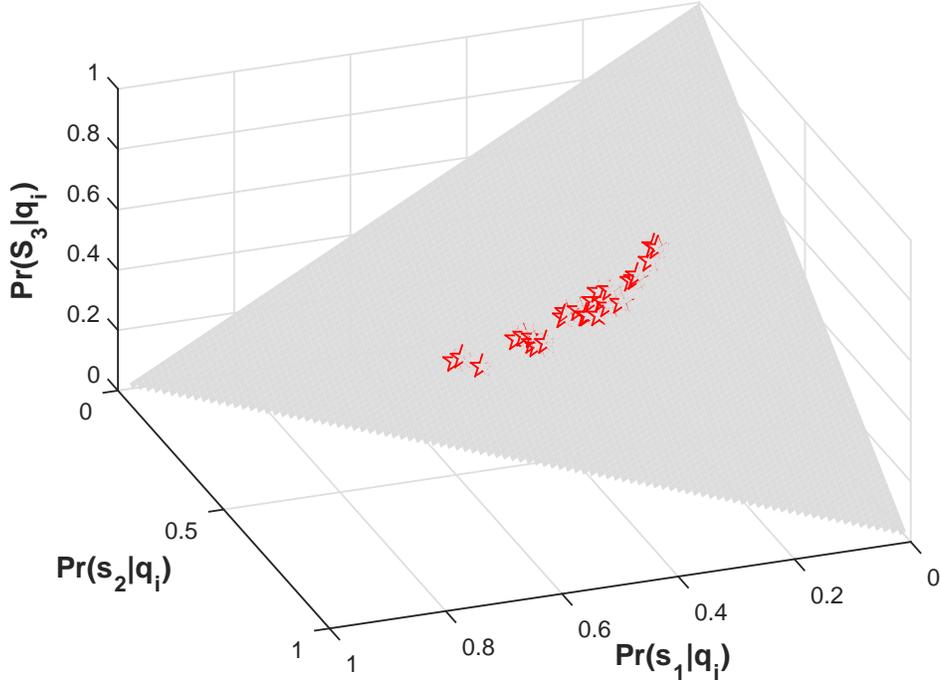}
	\caption{The symbol emission probabilities for a Markov chain with $3$ symbols are shown on a simplex. Symmetric K-L distance is used to find the structure in the state-set in the information space and the states are clustered based on the revealed structure.}
	\label{fig:Simplexcluster}
\end{figure}

The states of the reduced-order Markov model are then estimated by partitioning the set of words over $\alphabetSet$ of length $\depth$ estimated in the last step. This is done by using an agglomerative hierarchical clustering approach. The advantage of using the hierarchical clustering approach is that it helps visualize the structure of the set of the original states using an appropriate metric. Agglomerative hierarchical clustering is a bottom-up clustering approach~\cite{XW05} that generates a sparse network (e.g., a binary tree) of the state set $\stateSet$ (where $|Q|=|\alphabetSet|^\depth$) by successive addition of edges between the elements of $\stateSet$. Initially, each of the states $\state_1,\state_2,\dots,\state_n$  is in its own cluster $C_1,C_2,\dots, C_n$ where $C_i\in \mathcal{C}$, which is the set of all clusters for the hierarchical cluster tree. The distance between any two states in $\stateSet$ is measured using the K-L distance between the symbol emission probabilities conditioned on them, i.e.,
\begin{align}\label{eq:kldistance}
	d(\state_i,\state_j)& =  D_{\textrm{KL}}(\prob(\alphabetSet|\state_i)\|\prob(\alphabetSet|\state_j))\nonumber \\
	&+D_{\textrm{KL}}(\prob(\alphabetSet|\state_j)\|\prob(\alphabetSet|\state_i))
\end{align}
where the terms on the right have the following meaning.
\begin{align}
& D_{\textrm{KL}}(\prob(\alphabetSet|\state_i)\|\prob(\alphabetSet|\state_j)) \nonumber \\
& = \sum_{\symb \in \alphabetSet}\prob(\symb|q_i)\log\bigg( \frac{\prob(\symb|q_i)}{\prob(\symb|q_j)}\bigg) \nonumber
\end{align}
In terms of the distance measured by eq.~\eqref{eq:kldistance}, the pair of clusters that are nearest to each other are merged and this step is repeated till only one cluster is left. The tree structure displays the order of splits in the state set of the higher-order Markov model and is used to aggregate the states close to each other. For clarification of presentation, we show an example of a Markov chain with $27$ states and $3$ symbols on a simplex plane in Figure~\ref{fig:Simplexcluster}, where each \color{red} red pentagon \color{black} on the simplex represents one row of the symbol emission matrix. The hierarchical clustering is used to find the structure of the state set on the simplex place using the K-L distance. The set of states clustered together could be obtained based on the number of final states required in the final Markov model.

The overall algorithm is presented as a pseudo-code in Algorithm~\ref{algorithm:Modeling}. This algorithms is used to find the parameters of the models during training. The parameters during test are estimated using the clustering map $f_{N_{\textrm{max}}}$ and is further discussed in next section. In the later sections we show how an information theoretic criterion could be used to select the appropriate model to terminate the state merging algorithm or select a final model from the set of reduced-order models. Through numerical experiments using two different data-sets we also illustrate the main motivation of this work that although the right memory is required for accurate modeling of the symbolic process, the state-space not necessarily consist of all words corresponding to the estimated memory and we can achieve sufficiently-high predictive accuracy even with a smaller state-space. We are able to achieve this trade-off between the model complexity and predictive modeling accuracy using the information-theoretic criteria.

\begin{algorithm}[h] \small
\SetKwInput{KwIn}{Input}
\SetKwInput{KwOut}{Output}

	\KwIn {The observed symbol sequence $\symbSeq=\{\dots s_1s_2s_3\dots|s_i \in \alphabetSet\}$}% and the desired number of states in final model $N_{\textrm{max}}\leq \mid\stateSet\mid$}
	
	\KwOut {The final Markov model, $\mathcal{M}=(\tilde{\stateSet},\tilde{\emMat},\tilde{\trMat})$}	
	
	Estimate the $\trMat$ matrix for 1-step Markov model using frequency counting with an uniform prior\; %equation~\eqref{eqn:regressionEqn}\;
	
	Estimate the size of temporal memory, $\depth(\epsilon)$ for $\symbSeq$ using equation~\eqref{eq:depthTrace}\;
	
	Estimate $\emMat$ and $\trMat$ for the $\depth(\epsilon)$-Markov model using frequency counting with an uniform prior\;
	
	$\mathcal{C}_{\mid\stateSet\mid}=\{q_i\mid q_i \in \stateSet\}$\;
	
	\For{$i = \mid\stateSet\mid-1,\dots,1$}{{find distinct clusters $A,B \in \mathcal{C}_{i+1}$  minimizing $d(A\cup B)$\;}{$\mathcal{C}_i:=(\mathcal{C}_{i+1}\setminus\{A,B\})\cup \{A\cup B\}$}}
	\Return{$\mathcal{C}_1,\dots,\mathcal{C}_{\mid\stateSet\mid}$ and $f_i:\stateSet \rightarrow \mathcal{C}_i$} $\forall i \in \{1,\dots,\mid\stateSet\mid\}$
	
	Calculate the parameters of reduced model using $\tilde{\stateSet}=\mathcal{C}_{N_{\textrm{max}}}$, $f_{N_{\textrm{max}}}$ and equations~\eqref{eq:ParameterEst} through~\eqref{eq:pq1q2}\;
	
	Calculate the Log-likelihood for models with Equation~\eqref{eq:depth_loglikelihood}\;
	
	The final model is selected using the AIC or BIC criteria explained in Section~\ref{subsec:MDL}\;
	%Calculate the set $D_n$ where $D_n(x)=c_{n,\alpha_{\mathrm{CI}}}(x)$\;
	\caption{Reduced Order Markov Modeling}
	\label{algorithm:Modeling}
\end{algorithm}

\subsection{Parameter Estimation of the Reduced-Order Markov Model}\label{subsec:DBN}
The parameters of the Markov model obtained after clustering the states of the original PFSA with $|\alphabetSet|^\depth$ states is obtained using a Bayesian inference technique using the parameters estimated for the PFSA. In this proposed approach, the state transition matrix $\trMat$, the emission matrix $\emMat$, and the state probability vector $\stProbVec$ of the original PFSA model $\pfsa$ are available, along with the deterministic assignment map $f:\stateSet \rightarrow \widetilde{\stateSet}$ of the state in $\stateSet$ (i.e., state set of original model) to one of the state in $\widetilde{\stateSet}$ (i.e., state set of the reduced order model).
Since the reduced order model can represented by the tuple $\widetilde{\pfsa} = (\widetilde{\stateSet}, \widetilde{\trMat})$, where $\widetilde{\trMat} = [\tilde{\trProb}_{ij}]$ is the state transition matrix, we employ a Bayesian inference technique to infer the individual values of transition probabilities $\tilde{\trProb}_{ij} = \prob(\tilde{\state}_{k+1} = j \mid \tilde{\state}_{k} = i)$ for all $i, j \in \widetilde{\stateSet}$.

\begin{figure*} %Fig 03
	\centering
	\includegraphics[width=0.75\textwidth]{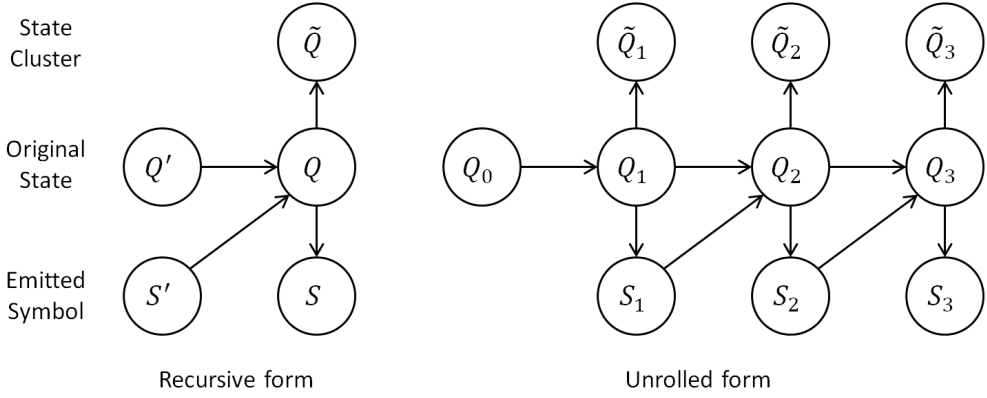}
	\caption{Graphical models representing the dependencies between the random variables}
	\label{fig:DBN}
\end{figure*}

Let $\stateVar_{k}$ be the random variable denoting the state of PFSA model at some time step $k \in \natural$ and $\symbVar_{k}$ denotes the symbol emitted from that state, this probabilistic emission process is governed by the emission matrix $\emMat$. The state of the reduced order model is obtained from a deterministic mapping of the state of the PFSA model, thus the state of this model is also a random variable, which is denoted by $\widetilde{\stateVar}_{k} = f(\stateVar_{k})$. The Bayesian network representing the dependencies between these variables is shown in the recursive as well as unrolled form in the Figure~\ref{fig:DBN}.
The conditional density $\prob(\widetilde{\stateVar}_{k} = \tilde{q} \mid \stateVar_{k} = q)$ can be evaluated by checking if state $q$ belongs to the state cluster $\tilde{q}$ and assigning the value of 1 if true, else assign it the value of 0. Since we know that $\widetilde{\stateSet}$ partitions the set $\stateSet$, the conditional density is well-defined. Thus, it can be written as
\begin{align}\label{eq:pc1q1}
	\prob(\widetilde{\stateVar}_{k} = \tilde{\state} \mid \stateVar_{k} = \state) = \indicator{\tilde{\state}}{\state},
\end{align}
where $\operatorname{I}$ is the indicator function with $\indicator{\tilde{q}}{q} = 1$, if element $q$ belongs to the set $\tilde{q}$, else it is $0$. The derivation of the Markov model $\prob(\widetilde{\stateVar}_{k+1}\mid \widetilde{\stateVar}_{k})$ using $\prob(\stateVar_{k+1}\mid \stateVar_{k})$, stationary probability vector $\stProbVec$, and assignment map $f$ is shown ahead.
\begin{align}\label{eq:ParameterEst}
	%\intertext{We write the Markov model using marginalization of the density $\prob(\widetilde{\stateVar}_{k+1}, \stateVar_{k+1} \mid \widetilde{\stateVar}_{k})$}
	&\prob(\widetilde{\stateVar}_{k+1}\mid \widetilde{\stateVar}_{k}) = \sum_{\state \in \stateSet} \prob(\widetilde{\stateVar}_{k+1}, \stateVar_{k+1} = q \mid \widetilde{\stateVar}_{k})  \\
	& \text{(Marginalization)}\notag\\
	%\intertext{Using chain rule of probability we get}
	%&= \sum_{\state \in \stateSet} \prob(\stateVar_{k+1} = q \mid \widetilde{\stateVar}_{k}) \prob(\widetilde{\stateVar}_{k+1} \mid \widetilde{\stateVar}_{k}, \stateVar_{k+1} = q) & \text{(Chain rule of probability)}\notag\\
	&\phantom{\prob(\widetilde{\stateVar}} = \sum_{\state \in \stateSet} \prob(\stateVar_{k+1} = q \mid \widetilde{\stateVar}_{k}) \prob(\widetilde{\stateVar}_{k+1} \mid \stateVar_{k+1} = q)\\
	& \text{(Factorization using Figure~\ref{fig:DBN})}\notag\\
	&\phantom{\prob(\widetilde{\stateVar}} = \sum_{\state \in \stateSet} \prob(\stateVar_{k+1} = q \mid \widetilde{\stateVar}_{k}) \indicator{\widetilde{\stateVar}_{k+1}}{\state} \\
	& \text{(using~\eqref{eq:pc1q1})}\notag\\
	&\phantom{\prob(\widetilde{\stateVar}} = \sum_{\state \in \widetilde{\stateVar}_{k+1}} \prob(\stateVar_{k+1} = q \mid \widetilde{\stateVar}_{k}) \label{eq:pc2c1}.
\end{align}
We can obtain $\prob(\stateVar_{k+1} \mid \widetilde{\stateVar}_{k})$ from Bayes' rule as
\begin{align}\label{eq:pq2c1}
	\prob(\stateVar_{k+1} \mid \widetilde{\stateVar}_{k}) = \dfrac{\prob(\widetilde{\stateVar}_{k} \mid \stateVar_{k+1})\prob(\stateVar_{k+1})}{\sum_{q \in \stateSet}\prob(\widetilde{\stateVar}_{k} \mid \stateVar_{k+1}=q)\prob(\stateVar_{k+1} = q)}.
\end{align}
Following the steps to obtain~\eqref{eq:pc2c1}, we also derive
\begin{align}\label{eq:pc1q2}
	\prob(\widetilde{\stateVar}_{k} \mid \stateVar_{k+1}) = \sum_{\state \in \widetilde{\stateVar}_{k}} \prob(\stateVar_{k} = q \mid \stateVar_{k+1}) .
\end{align}
We can obtain $\prob(\stateVar_{k} \mid \stateVar_{k+1})$ from Bayes' rule as
\begin{align}\label{eq:pq1q2}
	\prob(\stateVar_{k} \mid \stateVar_{k+1}) = \dfrac{\prob(\stateVar_{k+1} \mid \stateVar_{k})\prob(\stateVar_{k})}{\sum_{q \in \stateSet}\prob(\stateVar_{k+1} \mid \stateVar_{k}=q)\prob(\stateVar_{k} = q)}.
\end{align}
Note that, for the distribution $\prob(\stateVar_{k})$ and $\prob(\stateVar_{k+1})$, we use the stationary probability $\stProbVec$. Using the equations \eqref{eq:pc2c1}, \eqref{eq:pq2c1},
\eqref{eq:pc1q2}, and \eqref{eq:pq1q2} together, one can easily obtain the desired state transition matrix $\widetilde{\trMat}$ of the reduced order model. Once the state cluster set $\widetilde{\stateSet}$ and state transition matrix $\widetilde{\trMat}$ are available, the reduced order model is completely defined.

\subsection{Model Selection using information theoretic criteria}\label{subsec:MDL}
In this section, we describe the model selection process during the underlying state merging process for model inference. We compute ``penalized'' likelihood estimates for different models. Then, the model with the lowest score is selection as the optimal model.

The (unpenalized) log-likelihood of a symbol sequence $\symbSeq$ given a Markov model $\pfsa$ is computed as follows:
\begin{align}\label{eq:loglikelihood_Markov}
%\nonumber
\loglike(\symbSeq | \pfsa) \cong
\sum_{k=1}^{N}\log \prob\left( \symb_{k} | \state_k \right)
\end{align}
where the effects of the initial state are ignored because they become negligible for long statistically stationary symbol sequences. It is noted that with a finite symbol sequence, the log-likelihood is always finite. Furthermore, with the Markov models considered in this paper, the sum is simplified to the following form.
\begin{align}\label{eq:loglikelihood_DMarkov}
%\nonumber
\loglike(\symbSeq | \pfsa) \cong
\sum_{k=\depth+1}^{N}\log \prob\left( \symb_{k} | \symb_{k-1},\dots,\symb_{k-\depth} \right)
\end{align}

As discussed earlier, the states are merged using hierarchical clustering and thus, for every desired number of final states we get the deterministic map $f_{N_{\textrm{max}}}$ which determines how the original states are partitioned using the hierarchical clustering. This map is known for every terminal number of states and thus, we can find the log-likelihood of the symbol sequence using the following relationship.
\begin{align}\label{eq:depth_loglikelihood}
%\nonumber
\loglike(\symbSeq | \tilde{\pfsa}) \cong
\sum_{k=\depth+1}^{N}\log \prob\left( \symb_{k} | \tilde{\state}_k=f_{N_{\textrm{max}}}(\state_k) \right)
\end{align}
where, $\tilde{\state}_k$ is the state of the reduced model and $\state_k$ is the state of the original full-order model.

In the next step of the model selection process, a ``complexity penalty'' is added to the log-likelihood estimates, thereby balancing goodness of fit against the complexity of the model (and hence trying to prevent overfitting). We apply two widely-used such model selection functions, namely the Akaike information criterion (AIC)~\cite{Akaike:1974a} and the Bayesian information criterion (BIC)~\cite{Schwarz:1978a}:
\begin{enumerate}
\item $\mathcal{M}_{\textrm{BIC}}=-2\loglike(\symbSeq | \tilde{\pfsa})+K\log(N)$, where $K$ is the number of free parameters and $N$ is the number of observations.
\item $\mathcal{M}_{\textrm{AIC}}=-2\loglike(\symbSeq | \tilde{\pfsa})+2K$, where $K$ is the number of free parameters.
\end{enumerate}
The number of free parameters to be estimated from the data is the parameters of the symbol emission parameters, i.e., $K=\mid\alphabetSet\mid \mid\tilde{\stateSet}\mid$. It is noted that this allows model selection for individual symbol sequences. The criterion here allows a terminal condition for state merging; however, different symbol sequences can have different models. The model with the minimum score is selected as the best model. Through the results presented in next sections we illustrate the fact that most of the temporal and predictive capabilities can be preserved for the models with a very small number of states when compared to the original model.
\begin{remark}
	The final Markov model is a finite depth approximation of the original time-series data. However, compared to the PFSA-based D-Markov machines in~\cite{R04, MR14}, the current aggregated model has a non-deterministic algebraic structure, i.e., the same symbol emissions from a state can lead to different states. While this leads to some loss in predictive capability as compared to the models in~\cite{R04, MR14}, this allows us to compress the size of the model as per the requirement at hand. This allows faster convergence rates for the symbol emission probabilities as we only require fewer parameters to estimate from data, which might lead to faster decisions during testing.
\end{remark}
In the rest of the paper, we will present a Hamming distance-based bound for distortion in the predictive capabilities of reduced models and demonstrate the utility of these models in practical problems of fault/anomaly detection from time-series data.
\section{Analysis of the Proposed Algorithm}\label{sec:analysis}
In this section, we will present a bound on the distortion of the model due to the reduction of state-space of the Markov model using Hamming distance between two symbol sequences. We first present the Pinsker's inequality~\cite{G90} which relates the information divergence with the variational distance between probability measures defined on arbitrary spaces. This is followed by another theorem which can be used to derive Hamming distance bounds using the informational divergence.
\begin{theorem}[Pinsker's inequality]~\cite{G90}
Let $P$ and $Q$ be two probability distributions on a measurable space $(\mathds{X},\Sigma)$. Then, the following is true
\begin{eqnarray}
d_{TV}(P,Q)\leq \sqrt{\frac{1}{2}D_{\textrm {KL}}(P\|Q)}
\end{eqnarray}
where $d_{TV}(P,Q)=\sup\limits_{A\in \Sigma} \{|P(A)-Q(A)|\}$ is the total variation distance.
\end{theorem}
\begin{theorem}~\cite{M96}
Let $\mathds X$ be a countable set and let us denote by $x^n$ the sequence $(x_1,x_2,\dots,x_n)\in \mathds{X}^n$. Let $q^n$ be a Markov measure on $\mathds X^n$, that is, $q(x^n)=q(x_1)\prod\limits_{i=2}^n q_i(x_i|x_{i-1})$. Then for any probability measure $p^n$ on $\mathds X^n$, the following is true
\begin{eqnarray}
\bar{d}(p^n,q^n)\leq \bigg[\frac{1}{2n}D_{\textrm {KL}}(p^n\|q^n)\bigg]^{1/2}
\end{eqnarray}
where, $\bar{d}$ denotes the normed Hamming distance on $\mathds X^n \times \mathds X^n:$
\begin{equation}
\bar{d}(x^n,y^n)=n^{-1}\sum\limits_{n=1}^n d(x_i,y_i),
\end{equation}
where $d(x_i,y_i)=1$ if $x_i\neq y_i$ and $0$ otherwise. The $\bar{d}$-distance between $p^n$ and $q^n$ is
\begin{eqnarray}
\bar{d}(p^n,q^n)=\min E  \bar{d}(\hat{X}^n,X^n),
\end{eqnarray}
where $\min$ is taken over all joint distributions with marginals $p^n=\mathrm{dist} \hat{X}^n$ and $q^n=\mathrm{dist} {X}^n$ and $E$ denotes the expectation operator.
\end{theorem}

The above theorem provides us a way to bound Hamming distance between sequences generated by two different distributions. Thus, using the above theorem, we find a bound on the Hamming distance between the symbol sequences generated by the reduced-order Markov model and the original model by estimating the K-L distance between the measure on symbol sequences induced by these models. An approximate estimate of the K-L distance between the original and a reduced model could be expressed and estimated as shown in the following.

 Let the original D-Markov model be denoted by $\mathcal{M}$ and the reduced-order model by $\hat{\mathcal{M}}$. The Markov measure on the probability space $(S^n,\mathcal{E},P)$ where the set $S^n$ consists of sequences of length $n$ from an alphabet $\alphabetSet$ could be estimated using the symbol emission probabilities. More explicitly, the Markov measure of a sequence $S_n$ on $S^n$ induced by $\mathcal{M}$ is given by $P_{\mathcal{M}}(S_n)=\prob(\state_1)\prod\limits_{i=D+1}^n\prob(s_i\mid \state_i)$ (where $\depth$ is the depth of the model). Then, the K-L divergence between $\mathcal{M}$ and $\hat{\mathcal{M}}$ is given by the following expression.
\begin{equation}~\label{HammingBound}
D_{\rm {KL}} (P^n_{\mathcal{M}}\|P^n_{\hat{\mathcal{M}}})=\sum\limits_{S_n \in S^n}P_{\mathcal{M}}(S_n) \log\bigg(\frac{P_{\mathcal{M}}(S_n)}{P_{\hat{\mathcal{M}}}(S_n)}\bigg)
\end{equation}
Then, the above expression can be simplified as follows.
\begin{equation}
\log\bigg(\frac{P_{\mathcal{M}}(S_n)}{P_{\hat{\mathcal{M}}}(S_n)}\bigg)=\sum\limits_{i=D+1}^n\log(\prob(s_i\mid q_i))-\log(\prob(s_i\mid \hat{q}_i)), \nonumber
\end{equation}
where, $\hat{q}$ is the merged state and $q$ is the original state. Then the expression on the right could be further bounded using the Lipschitz constant for the logarithm function and under the assumption that $\log(\prob(s_j\mid q_i))\neq 0$ $\forall q_i \in \stateSet$ and all $s_j \in \alphabetSet$.
\begin{align}
&\sum\limits_{i=D+1}^n\log(\prob(s_i\mid q_i))-\log(\prob(s_i\mid \hat{q}_i)) \label{eqn:logsum}\\
&\leq \sum\limits_{i=D+1}^n(\frac{\prob(s_i\mid q_i)-\prob(s_i\mid \hat{q}_i)}{\prob(s_i\mid q_i)})\label{eqn:lipschitz}\\
&\leq (n-D-1)\kappa
\end{align}
where, $\kappa=\max\limits_{q\in Q, s\in \alphabetSet}\frac{\prob(s\mid q)-\prob(s\mid \hat{q})}{\prob(s\mid q)}$. In the above inequalities, equation~\eqref{eqn:lipschitz} is obtained from equation~\eqref{eqn:logsum} by using the observation that $\prob(s_i\mid \hat{q}_i)=\prob(s_i\mid {q}_i)+\eta$, where $\eta$ is the perturbation in the symbol emission probability from $q_i$ when it is clustered into a new state $\hat{q}_i$. Hence, the K-L distance in equation~\eqref{HammingBound} could be bounded by the following term.
\begin{align}
D_{\rm {KL}} (P^n_{\mathcal{M}}\|P^n_{\hat{\mathcal{M}}})& \leq \sum\limits_{S_n \in S^n} P_{\mathcal{M}}(S_n) (n-D-1)\kappa \nonumber \\
& = (n-D-1)\kappa \sum\limits_{S_n \in S^n} P_{\mathcal{M}}(S_n) \nonumber \\
& =(n-D-1)\kappa
\end{align}
 Thus, a uniform bound on the Hamming distance between the original and the final model could then be obtained as follows,
\begin{equation}\label{eqn:bound}
\bar{d}(P_\mathcal{M}(S_n),P_{\mathcal{\hat{M}}}(S_n))\leq \sqrt{\frac{(n-D-1)\kappa}{2n}}
\end{equation}
The above inequality thus, allows us to compare models with different state-space based on the predictive accuracy of a reduced model when compared to the original model. As compared to the earlier information theoretic criteria, which were based on the efficiency of data compression by different models, the inequality in~\eqref{eqn:bound} allows to compare them based on their symbol emission statistics and thus, is computationally efficient. It is possible to find a rather tighter bound in an expected sense by using the stationary distribution of the two Markov chains to find an expected bound on Hamming distance. However, finding the same is left as an exercise for future work. Using the above bound for selection of models could be more efficient than the information theoretic metrics (as it can estimated by using the symbol emission probabilities instead of the penalized likelihoods); however, finding a penalized version of the bound for model selection is also left as a future exercise.

\section{Description of Experimentation and Data Sets}\label{sec:experiment}
In this section, we  briefly describe the two different data-sets which have been used in this paper to illustrate and validate the proposed concepts. Specifically, we will describe the experiments done at Penn State to investigate instability in lean-premixed combustion and another benchmark data-set for anomaly detection in bearings. An important point to be noted here is that the numerical experiments we present in the following sections is to justify the fact that the reduced-order models obtained by the proposed algorithms are able to achieve the trade-off between predictive accuracy and model complexity. Further results for classification and anomaly detection are to illustrate that this proposed approach of model learning can still achieve good performance for machine learning objectives of class separability and anomaly detection.
\subsection{Combustion}\label{subsec:combustionexperiment}
A swirl-stabilized, lean-premixed, laboratory-scale combustor was used to perform the experimental study. Tests were conducted at a nominal combustor pressure of 1 atm over a range of operating conditions, as listed in Table~\ref{tab:par}.

\begin{table}[!hbp]
	\centering
	\caption{Operating conditions}
	\begin{tabular}{c|c}
		\hline
		\textbf{Parameters} & \textbf{Value} \\
		\hline
		Equivalence Ratio & 0.525, 0.55, 0.60, 0.65\\
		\hline
		Inlet Velocity & 25-50 m/s i m/s increments \\
		\hline
		Combustor Length & 25-59 inch in 1 inch increments\\
		\hline
	\end{tabular}
	\label{tab:par}
\end{table}

 In each test, the combustion chamber dynamic pressure and the global OH and CH chemiluminescence intensity were measured to study the mechanisms of combustion instability. The measurements were made simultaneously at a sampling rate of 8192 Hz~(per channel), and data were collected for 8 seconds, for a total of 65536 measurements~(per channel). A total of $780$ samples of data were collected from all the tests where in every test the combustion process was driven from stable to unstable by changing either the equivalence ratio, $\phi$. However, as the accurate model of the process is not available, an accurate label of transition of the process to unstable phase is not available. It is noted that the data consists the behavior of the process over a large number of operating condition and thus provides a rich set of data to test the efficacy of the algorithm in detecting classes irrespective of the underlying operating conditions.
 \subsection{Bearing Prognostic Data} This test data has been picked from NASA's prognostics data repository~\cite{NASAPHM, TMZT12}. A detailed description of the experiments could be found in~\cite{QLLY06}. The bearing test rig hosts four test bearings on one shaft which is driven by an AC motor at a constant speed. A constant force is applied on each of the bearings and accelerometer data is collected at every bearing at a sampling rate of \SI{20}{\kilo\hertz} for about \SI{1}{\second}. The tests are carried for $35$ days until a significant amount of debris was found in the magnetic plug of the test bearing. A defect in at least one of the bearings is found at the end of every test. In this paper, we will use the data from a bearing which shows anomalous behavior in the later parts of test. In particular, out of the three data sets, we use set one where an inner race fault occurred on Bearing $3$. In the analysis, we use data from Bearing $3$.
\section{Markov Modeling}
\begin{figure}%Fig 04
	\centering \vspace{-6pt}
	\includegraphics[width=0.75\textwidth]{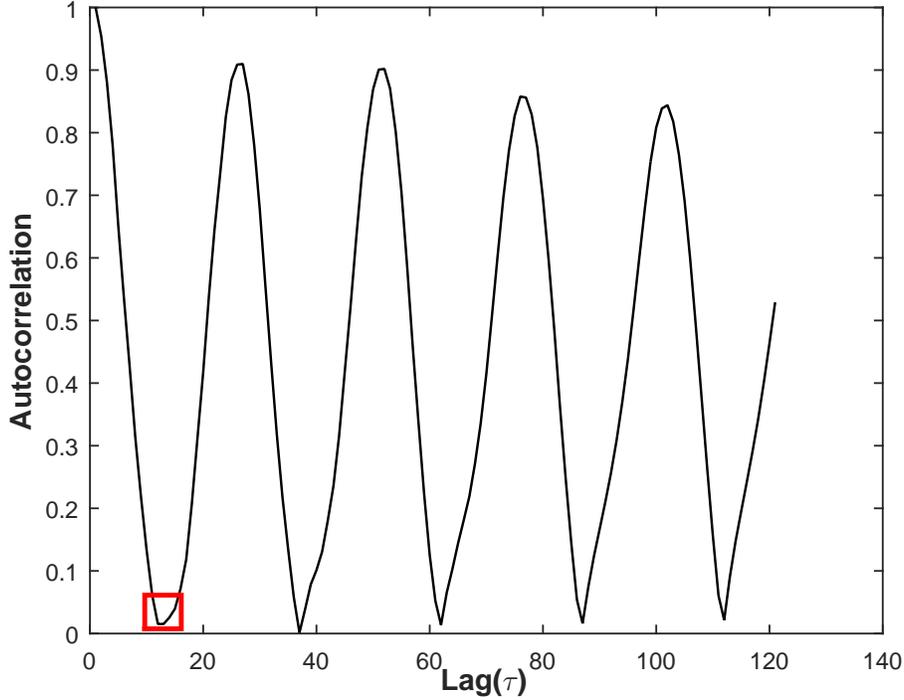}
	\caption{Autocorrelation function of time-series data during the unstable phase of combustion. The time-series data is down-sampled by the lag marked in red square. It is noted that the individual time-series have their own down-sampling lags.}
	\label{fig:autocorr}
\end{figure}

\begin{figure*} %Fig05
	\centering
	\subfloat[Probability density function for the pressure time series data]{\includegraphics[width=0.75\textwidth]{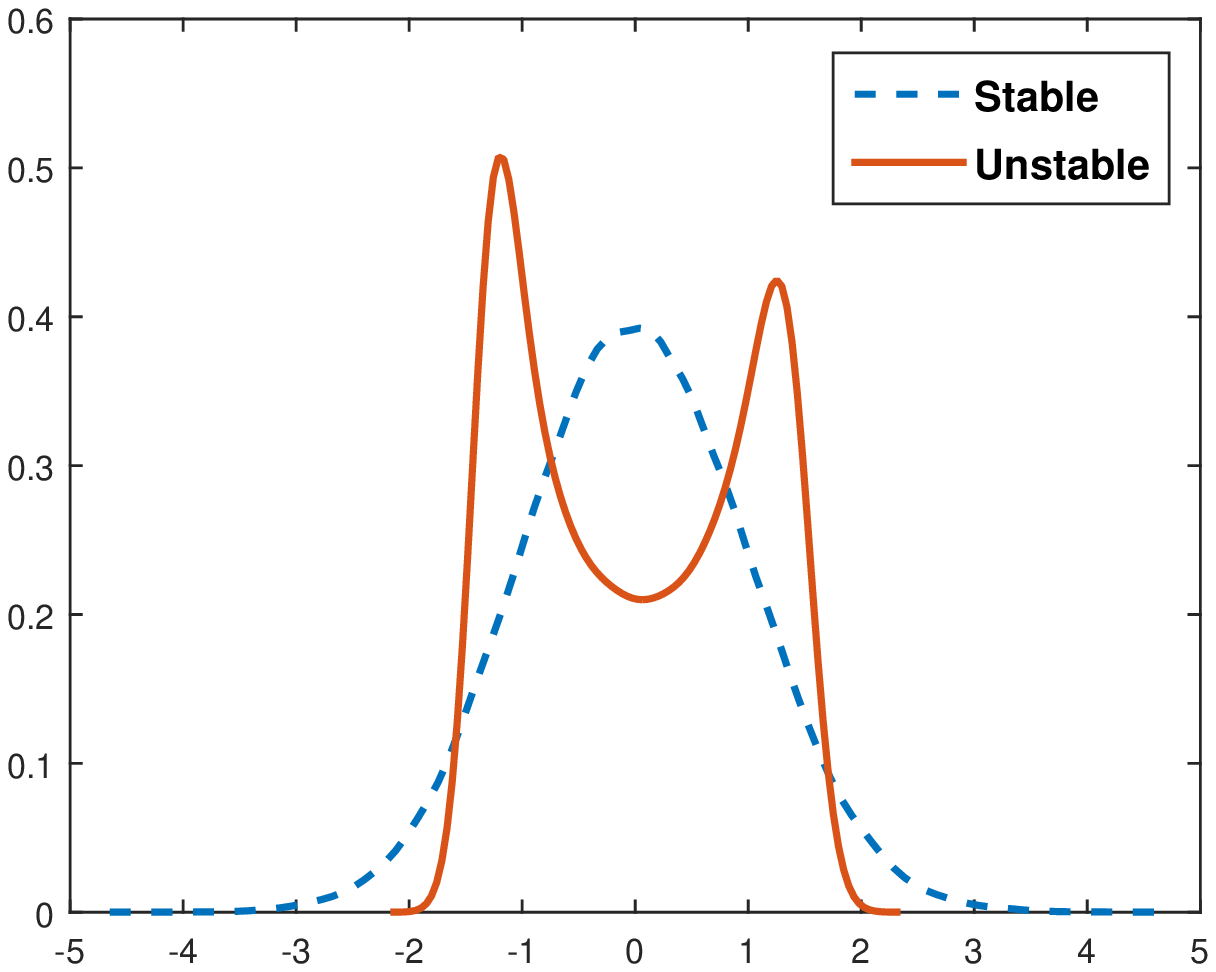}\label{fig:datadistribution}}\quad
	\subfloat[Spectral decomposition of the stochastic matrix for 1-step Markov model]{\includegraphics[width=0.75\textwidth]{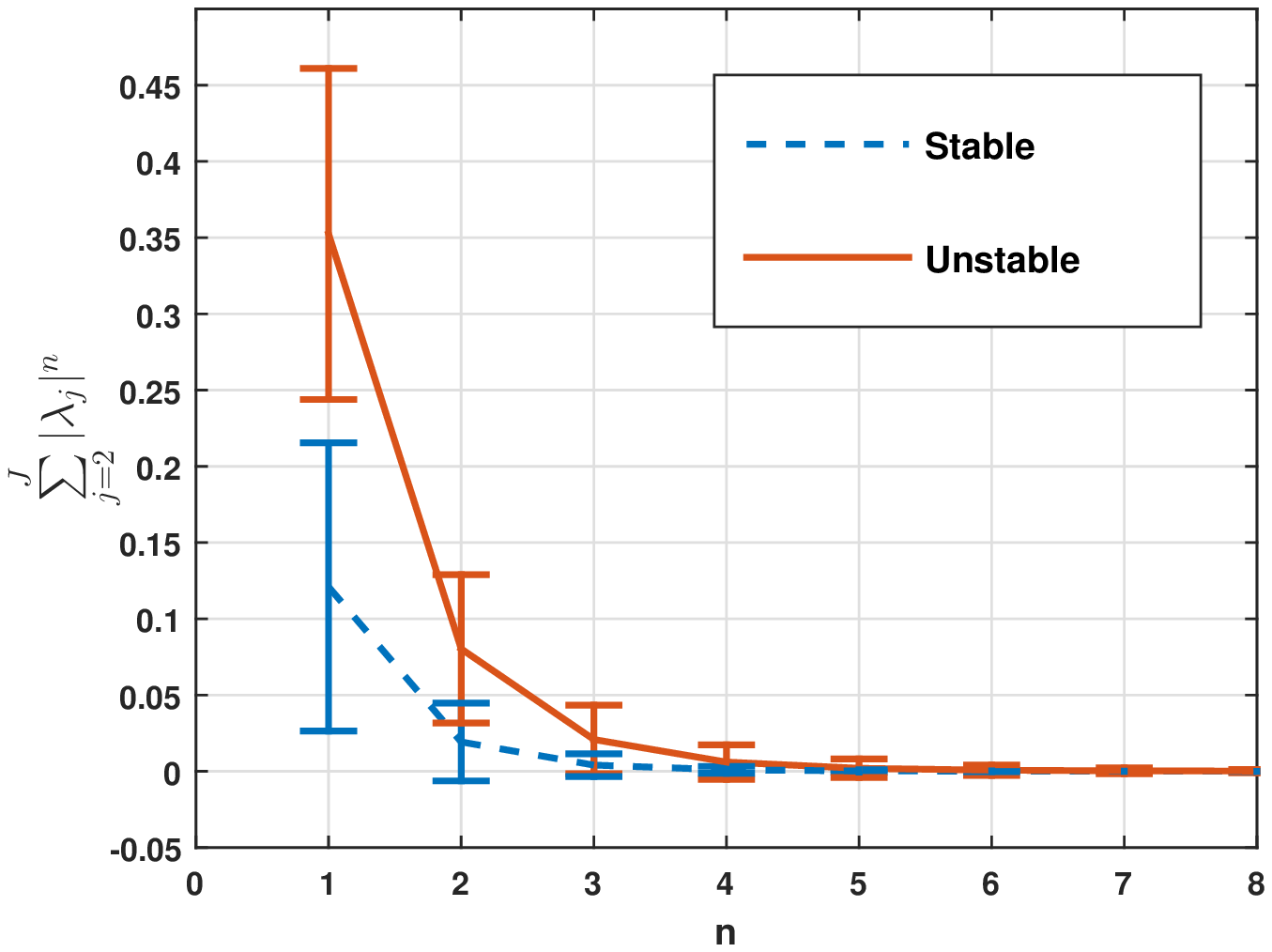}\label{fig:spectralprop}}\\
	\caption{The first plate in the above Figure shows the change in the empirical density calculated for the pressure time-series data as the process deviates from the stable operating condition to unstable operating condition. The second plate shows the spectral decomposition of the 1-step stochastic matrix for the data under stable and unstable operating conditions.}
	\label{fig:databehavior}\vspace{-2pt}
\end{figure*}

In this section, we present results for modeling and analysis of the time-series data which are presented in this paper.
\subsection{Combustion}
Time-series data is first normalized by subtracting the mean and dividing by the
standard deviation of its elements; this step corresponds to bias removal and
variance normalization. Data from engineering systems is typically oversampled
to ensure that the underlying dynamics can be captured (in the current experiments, it was $\SI{8192}{\hertz}$ ). Due to coarse-graining from the symbolization process, an over-sampled
time-series may mask the true nature of the system dynamics in the symbolic
domain (e.g., occurrence of self loops and irrelevant spurious transitions in
the Markov chain). Time-series is first down-sampled to find the next crucial
observation. The first minimum of auto-correlation function generated from the
observed time-series is obtained to find the uncorrelated samples in time. The
data sets are then down-sampled by this lag. The autocorrelation function for the time-series data during unstable case is shown in Figure~\ref{fig:autocorr} where the data are downsampled by the lag marked in \color{red} red \color{black} rectangle in Figure~\ref{fig:autocorr}. To avoid discarding significant amount of data due to downsampling, down-sampled data using different initial
conditions is concatenated. Further details of this preprocessing can be found
in~\cite{Srivastav2014}.

The continuous time-series data set is then partitioned using maximum entropy
partitioning (MEP), where the information rich regions of the data
set are partitioned finer and those with sparse information are partitioned
coarser. In essence, each cell in the partitioned data set contains
(approximately) equal number of data points under MEP. A ternary alphabet with
$\alphabetSet=\{0,1,2\}$ has been used to symbolize the continuous combustion
instability data. As discussed in section~\ref{sec:experiment}, we analyze data sets from different phases, as the process goes from stable through the transient to the unstable region (the ground truth is decided using the RMS-values of pressure).

\begin{figure} %Fig06
	\centering
	\subfloat[Hierarchical cluster tree of stable states]{\includegraphics[width=0.75\textwidth]{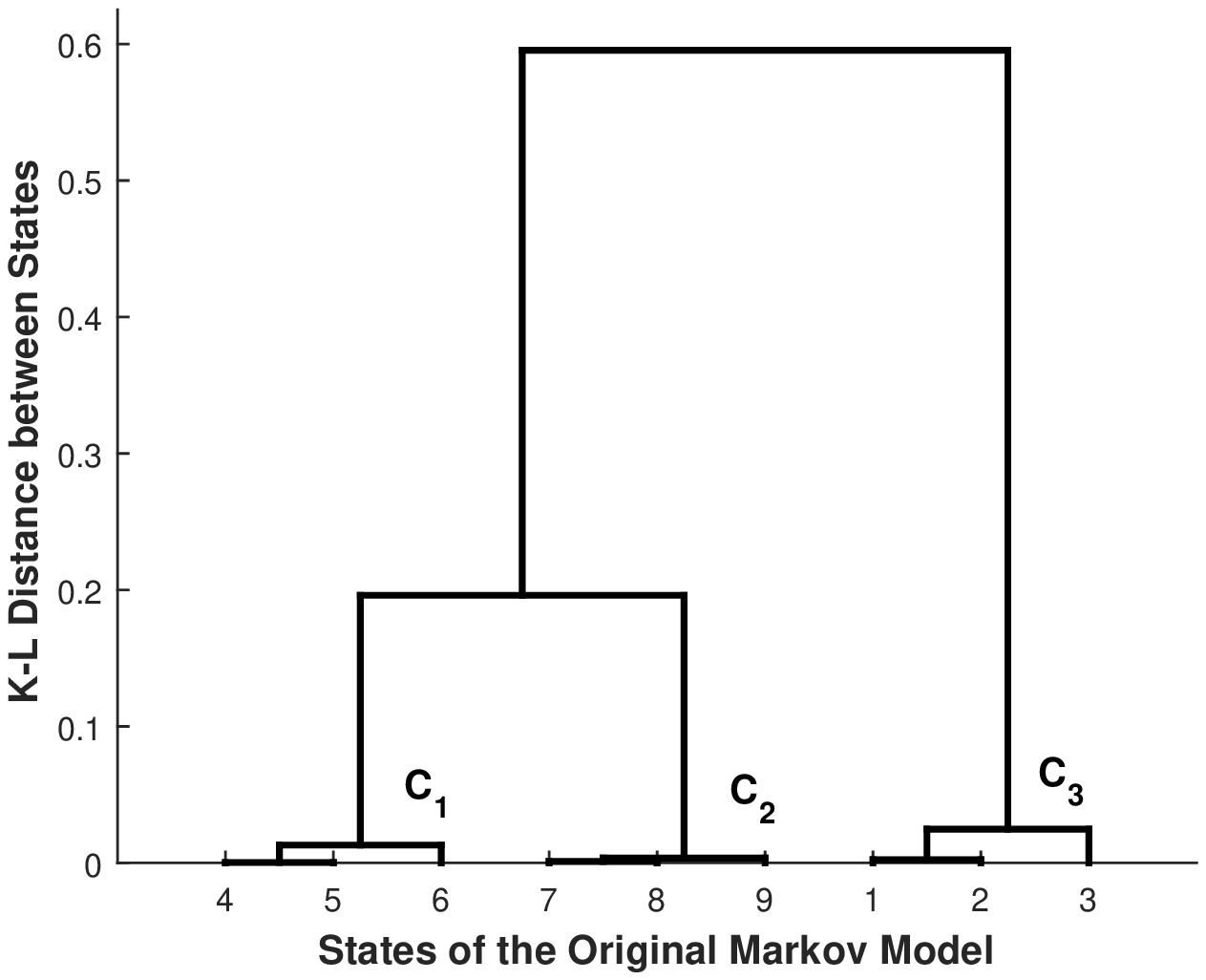}\label{fig:clusterstable}}\quad
	\subfloat[Hierarchical cluster tree of unstable states]{\includegraphics[width=0.75\textwidth]{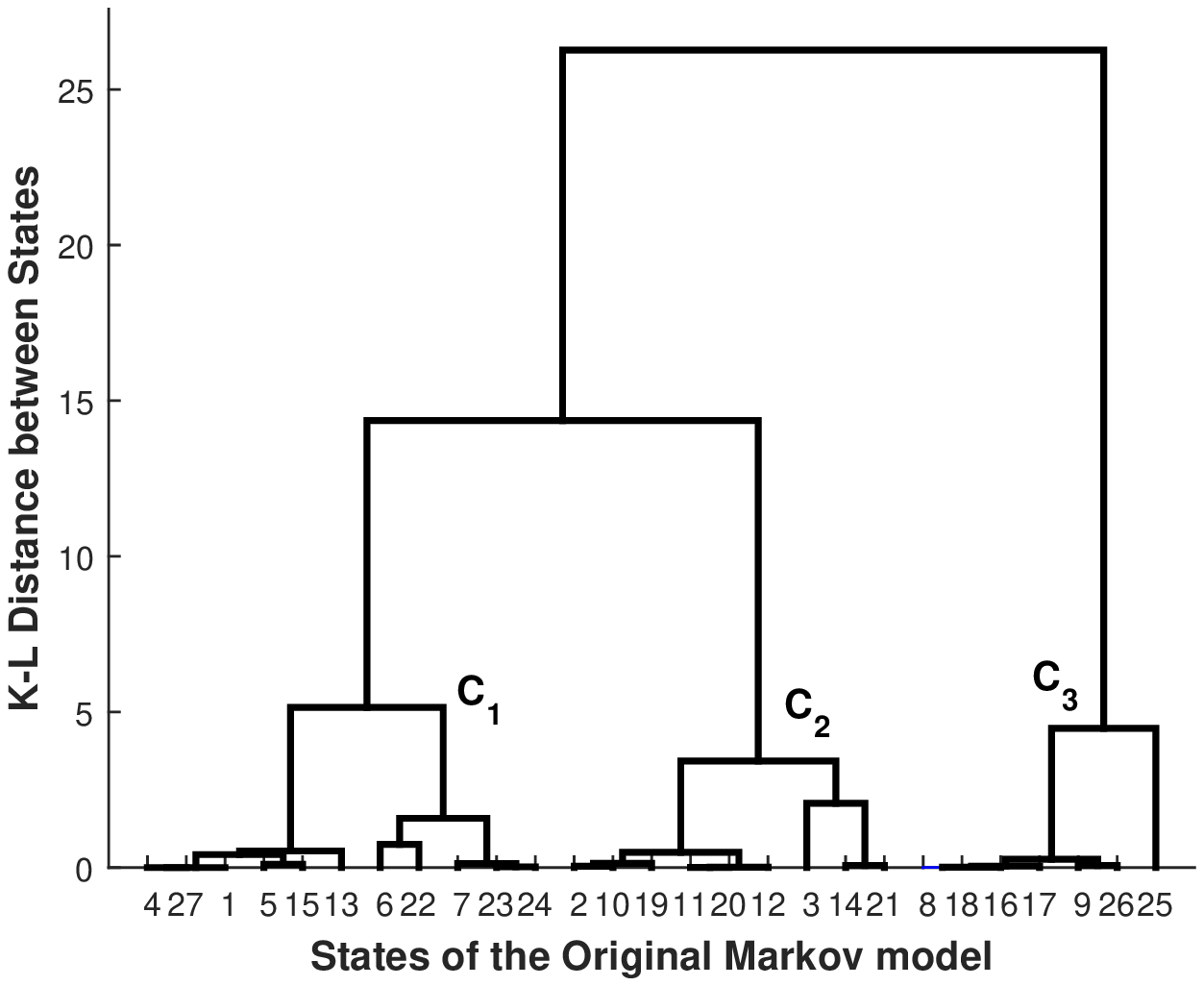}\label{fig:culsterunstable}}\\
	\caption{State clustering under stable and unstable conditions.}
	\label{fig:clusterbehavior}\vspace{-2pt}
\end{figure}

In Figure~\ref{fig:datadistribution}, we show the observed changes in the behavior of the data as the combustion operating condition changes from stable to unstable. A change in the empirical distribution of data from unimodal to bi-modal is observed as the system moves from stable to unstable. We selected $150$ samples of pressure data from the stable and unstable phases each to analyze and compare. First, we compare the expected size of temporal memory during the two stages of operation. There are changes in the eigenvalue decomposition rate for the 1-step stochastic matrix calculated from the data during the stable and unstable behavior, irrespective of the combustor length and inlet velocity. During stable conditions, the eigenvalues very quickly go to zero as compared to the unstable operating condition (see Figure~\ref{fig:spectralprop}). This suggests that the size of temporal memory of the discretized data increases as we move to the unstable operating condition. This indicates that under the stable operating condition, the discretized data behaves as symbolic noise as the predictive power of Markov models remain unaffected even if we increase the order of the Markov model. On the other hand, the predictive power of the Markov models can be increased by increasing the order of the Markov model during unstable operating condition, indicating more deterministic behavior. An $\epsilon=0.05$ is chosen to estimate the depth of the Markov models for both the stable and unstable phases. Correspondingly, the depth was calculated as $2$ and $3$ for the stable and unstable conditions (see Figure~\ref{fig:databehavior}). The corresponding $D(\epsilon)$ is used to construct the Markov models next. First a PFSA whose states are words over $\alphabetSet$ of length $\depth(\epsilon)$ is created and the corresponding maximum-likely parameters ($\emMat$ and $\trMat$) are estimated. Then, the hierarchical clustering algorithm using K-L distance is used to cluster and aggregate the states. It is noted that we create individual models for every sample, i.e., every sample is partitioned individually so that the symbols will have different meaning (i.e., they represent different regions in the measurement space of the signals) for every sample. Consequently, each sample will have a different state-space when viewed in the continuous domain. Thus, we do not show the mean behavior of the samples during any operating regime as the state-space would be inconsistent (even though the cardinality could be the same).

In Figure~\ref{fig:clusterbehavior}, we show the hierarchical cluster tree which details the structure of the state-space for the PFSA with depth $\depth(\epsilon)$ for a typical sample during stable and unstable behavior. The cluster tree also suggests the symbolic noise behavior of the data during the stable regime (the states are very close to each other based on the K-L distance). However, clearly a coarse clustering of states in the model during the unstable behavior would lead to significant information loss (as the states are statistically different). However, to compare the two Markov models, we keep the cardinality of the final models the same. For example, the algorithm is terminated with $3$ states in the final Markov model during the stable as well as the unstable regime. and the final aggregated states are the three clusters depicted in the Figure~\ref{fig:clusterbehavior}. Once the final aggregated states are obtained, we estimate the parameters of the model using the Bayesian inference discussed in section~\ref{subsec:DBN}.

\begin{figure} %Fig07
	\centering
	\subfloat[Model scores using the BIC and AIC criterion during a typical stable condition]{\includegraphics[width=0.75\textwidth]{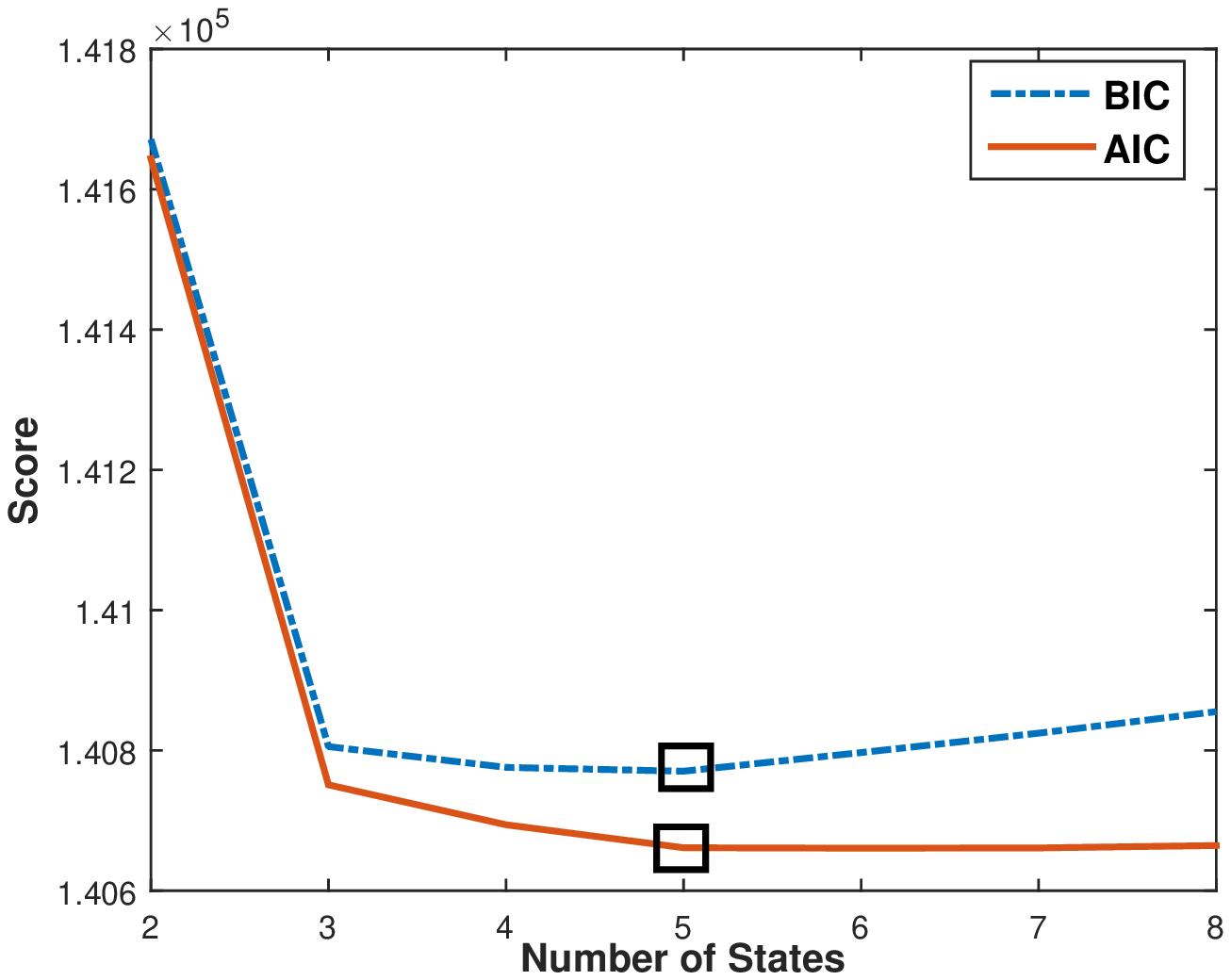}\label{fig:MDLstable}}\quad
	\subfloat[Model scores using the BIC and AIC criterion during a typical unstable condition]{\includegraphics[width=0.75\textwidth]{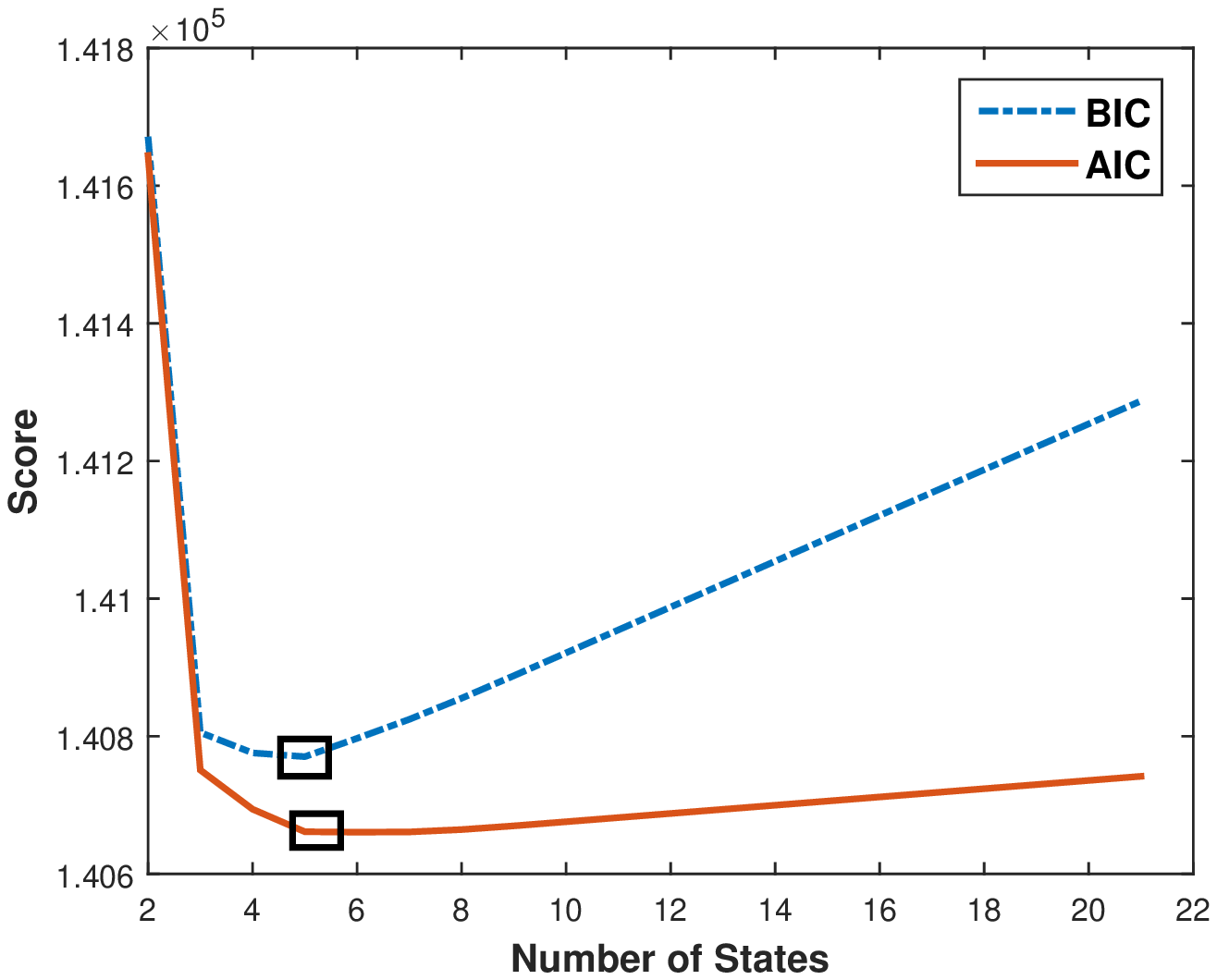}\label{fig:MDLunstable}}\\
	\caption{Unsupervised model selection under stable and unstable conditions.}
	\label{fig:MDLscores}\vspace{-2pt}
\end{figure}

\begin{figure} %Fig08
	\centering
	\subfloat[Hamming distance between the original and final models for a typical stable combustion sample]{\includegraphics[width=0.75\textwidth]{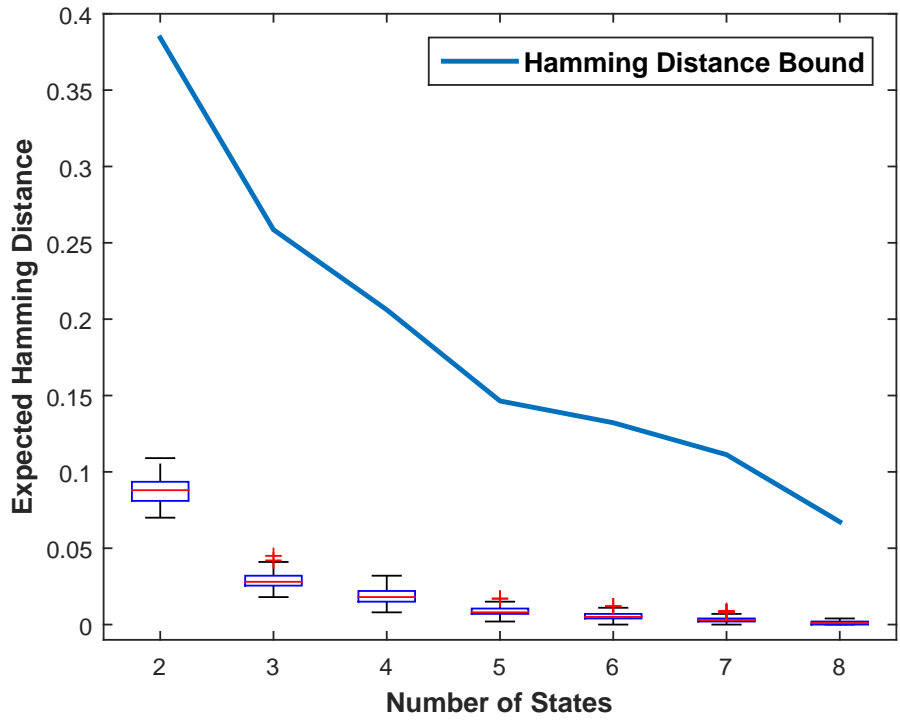}\label{fig:Hammingstable}}\quad
	\subfloat[Hamming distance between the original and final models for a typical unstable combustion sample]{\includegraphics[width=0.75\textwidth]{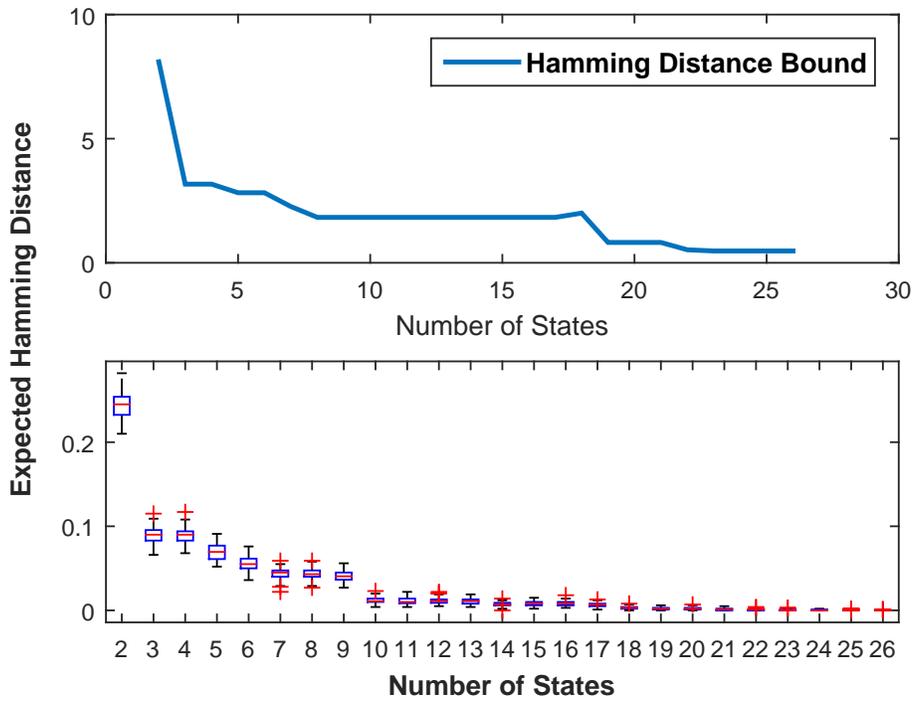}\label{fig:Hammingunstable}}\\
	\caption{Box plot for Hamming distance between the original and reduced-order models obtained after merging based on the results in Section~\ref{sec:analysis}}
	\label{fig:HammingCombustion}\vspace{-2pt}
\end{figure}

Next, we present some results for model selection using the information-theoretic criteria discussed earlier in section~\ref{subsec:MDL}. BIC and AIC are used to select the model which achieves the minimum score. As seen in the Figures~\ref{fig:MDLstable} through~\ref{fig:MDLunstable}, the model with $5$ states is selected for stable as well as for the unstable case (note that the original model for the stable class had $9$ states for depth $2$ and the unstable model had $27$ states for a depth of $3$). In contrast to cross-validation, the two criteria provide an unsupervised way for model selection. Thus we see that we need much smaller state-space to preserve the temporal statistics of the data and AIC and BIC provide us with a technique to select the compact model.
In Figure~\ref{fig:HammingCombustion}, we show the Hamming distance between the sequences generated by the original model and the reduced models for a typical sample each from stable and unstable combustion. The box-plots are generated by simulating the original model and the reduced-order model to generate symbol sequences of length $1000$ from $100$ different initial states (i.e., a total of $100$ strings are generated) and the Hamming distance between them is calculated. A bound on the Hamming distance between the sequences generated by the original model and final model is also calculated using the inequality~\eqref{eqn:bound}. The results are shown in Figure~\ref{fig:HammingCombustion}. It is possible to use the proposed Hamming distance metric to select a final model; however, this measures the distance between the distributions induced by the Markov models, and model selection using it is left as a future work. It is noted that the bounds on Hamming distance can provide a computationally convenient way to select model scores as it can be found from the symbol emission probabilities of the model instead of explicitly looking at the predictive capability by looking at the likelihoods of the symbol sequences.

\subsection{Bearing}
\begin{figure} %Fig09
	\centering
	\includegraphics[width=0.75\textwidth]{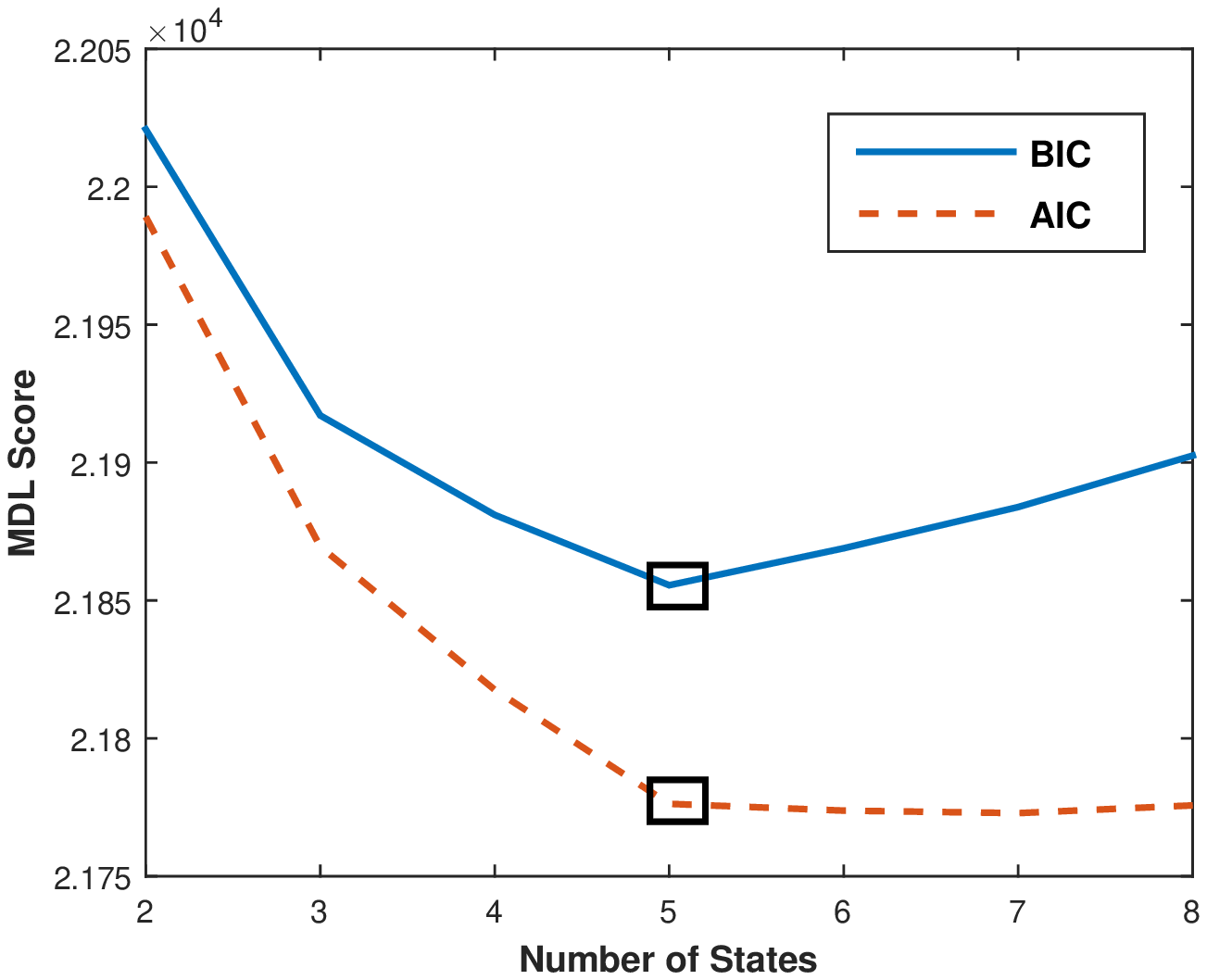}
	\caption{Model scores using the BIC and AIC criteria; selected models are depicted by black rectangles.}
	\label{fig:ModelBearing}
\end{figure}

\begin{figure} %Fig10
	\centering
	\includegraphics[width=0.75\textwidth]{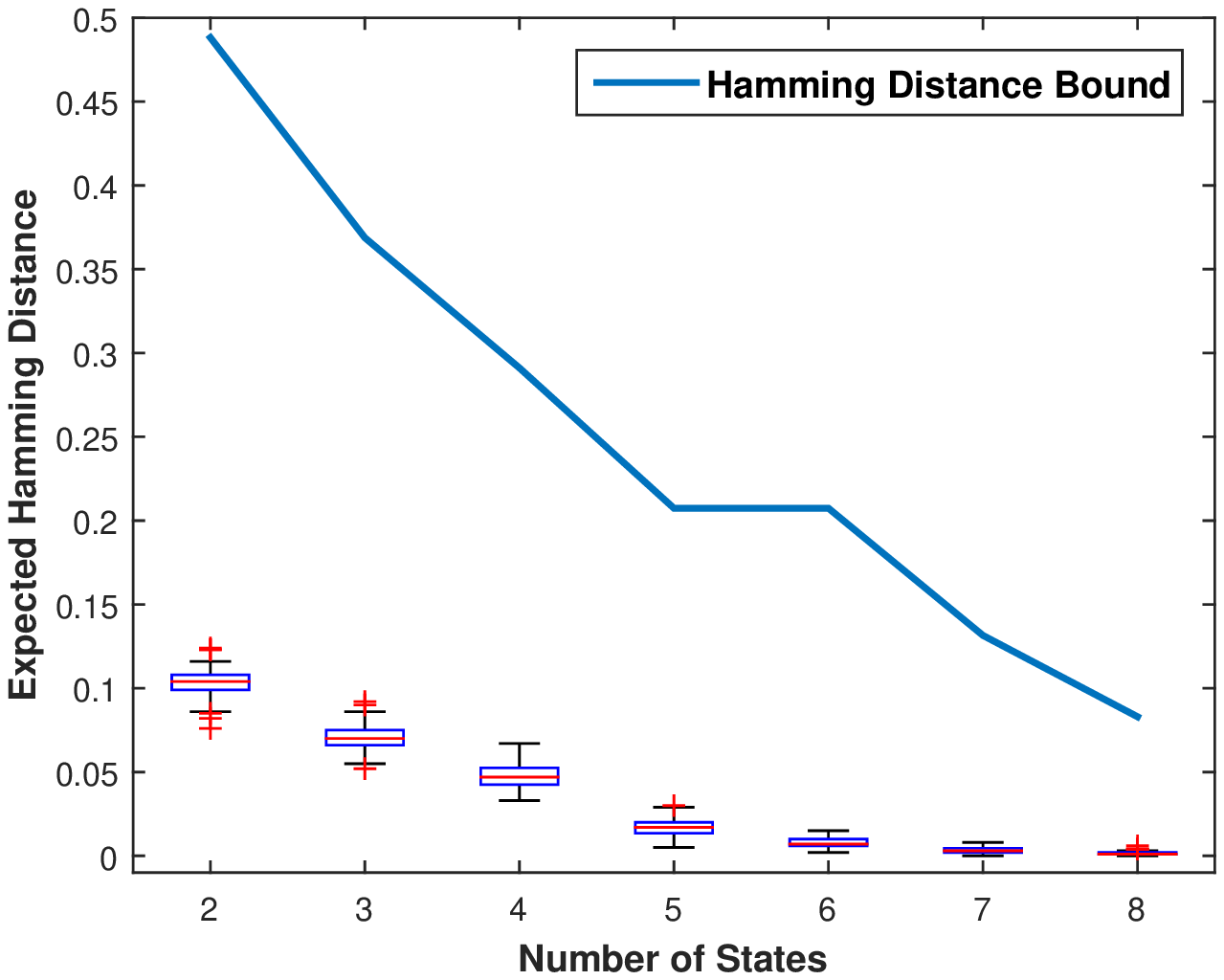}
	\caption{Box plot of the Hamming distance between the original and reduced-order models along with the analytical bound presented in Section~\ref{sec:analysis}.}
	\label{fig:HammingDistBearing}
\end{figure}

The same procedure of downsampling and depth estimation is followed for analysis of bearing data as was described in the previous section for combustion. A ternary alphabet is again chosen to discretize the continuous data after downsampling and the maximum entropy partitioning is used to find the partitions. Using the spectral method, a depth of $2$ (i.e., a total of $9$ states) is estimated for an $\epsilon=0.02$ (we skip the plot of spectral decomposition plot for brevity). The BIC and AIC score for the different models is shown in Figure~\ref{fig:ModelBearing} and the model with five states is selected using the obtained scores (marked in black rectangle). In Figure~\ref{fig:HammingDistBearing}, we show the Hamming distance between the sequences generated by the original model (with $9$ states) and the reduced models and the corresponding bounds obtained by inequality~\eqref{eqn:bound}.

\section{Classification and Anomaly Detection Results}
In this section, we present some results for anomaly detection and classification using the pressure time-series data to infer the underlying reduced-order Markov model. As we discussed earlier in section~\ref{subsec:combustionexperiment}, the exact transition point of the system from stable to unstable is unknown, we first present results on anomaly detection and clustering of the data into different clusters which can be then associated with the stable and unstable class. We will present two different metrics for anomaly detection that allows models of different state-space and structure to be compared. It is noted that the word metric is used here in a loose sense; it is meant to be a distance that could be used to compare two different Markov models.

\subsection{Anomaly Detection}
As individual time-series have different state-space, we define some metrics to compare them. These metrics reflect changes in the information complexity of Markov models and reveal different behavior of combustion process based on the changes in the inferred data model. In particular, the following two metrics are defined.
\begin{enumerate}
\item Cluster Divergence: This measure is defined for individual Markov models based on the cluster structure of the state-space of the model. Physically, it represents the maximum statistical difference between the states of the Markov model measures using K-L distance. It is calculated for a particular model $\mathcal{M}$ as follows
\begin{equation}\label{eqn:metric}
\Delta_{\mathcal{M}}=\max\limits_{q_i,q_j \in \stateSet} d(q_i,q_j)
\end{equation}
where $d$ is defined by equation~\eqref{eq:kldistance}.
\item Discrepancy Statistics: We measure the discrepancy between the i.i.d. statistics and the Markov statistics for the discretized data. This could be also interpreted as the information gain for Markov models. This measure also represents the information complexity of the data. If the i.i.d. statistics and the Markov statistics are very close, then the data has no temporal statistics; however, an increase in this measure would indicate the information gain by creating a temporal Markov model for the data. This is measured by the following equation.
\begin{equation}
H_\mathcal{M}=\sum_{q \in \stateSet} \prob(q) D_{KL}(\prob(\alphabetSet\mid q)\| \prob(\alphabetSet))
\end{equation}
where $\prob(\alphabetSet\mid q)$ represents the symbol emission probability conditioned on a state $q$ of the Markov model and $\prob(\alphabetSet)$ represents the marginal symbol emission probability. The term $D_{KL}$ represents the symmetric K-L distance between the two distributions.
\end{enumerate}
In Figure~\ref{fig:divergence}, we present some results to show the behavior of $\Delta_{\mathcal{M}}$ with increasing pressure fluctuations. It is noted that every model has been created in an unsupervised fashion by first discretizing and then, estimating the memory of the discrete sequence. As seen in Figure~\ref{fig:complexorig}, there are three distinct behaviors that can be associated with $\Delta_\mathcal{M}$. With low pressure fluctuations, the metric is very close to $0$, indicating that the states of the model are very similar statistically. This is seen until data number $200$ with corresponding $P_{rms}\sim 0.065 $ psig, which leads to a gradual change to a point where the measure saturates with $P_{rms}\sim 0.12$ psig (when the process becomes unstable). Thus, with this gradual trend with increasing pressure fluctuations, we associate different behaviors with the process. However, as is seen in the Figure~\ref{fig:complexorig}, the transition from stable to unstable behavior is not clearly defined and is very difficult to label during the experiments as the process is very fast. We show the pressure signals from the three different clusters in Figure~\ref{fig:pressure} where it could be seen that the sample number $250$ could be seen to approach an approximate limit cyclic behavior (and thus, could be loosely classified as transient stage).  An important point to note at this point is that this measure is independent of any operating conditions and only depends on stability (or instability) of the process. This metric is thus used for anomaly detection. In Figure~\ref{fig:complexfinal}, we show the statistics of $\Delta_{\mathcal M}$ with four states. We see that there is some loss of information up on merging states in the unstable class; the stable cluster remains unchanged implying that the states are statistically similar and the model distortion up on merging of states is insignificant. Thus, $\Delta_{\mathcal M}$ can be reliably used to detect departure from stable behavior.

\begin{figure} %Fig12
	\centering
	\includegraphics[width=0.75\textwidth]{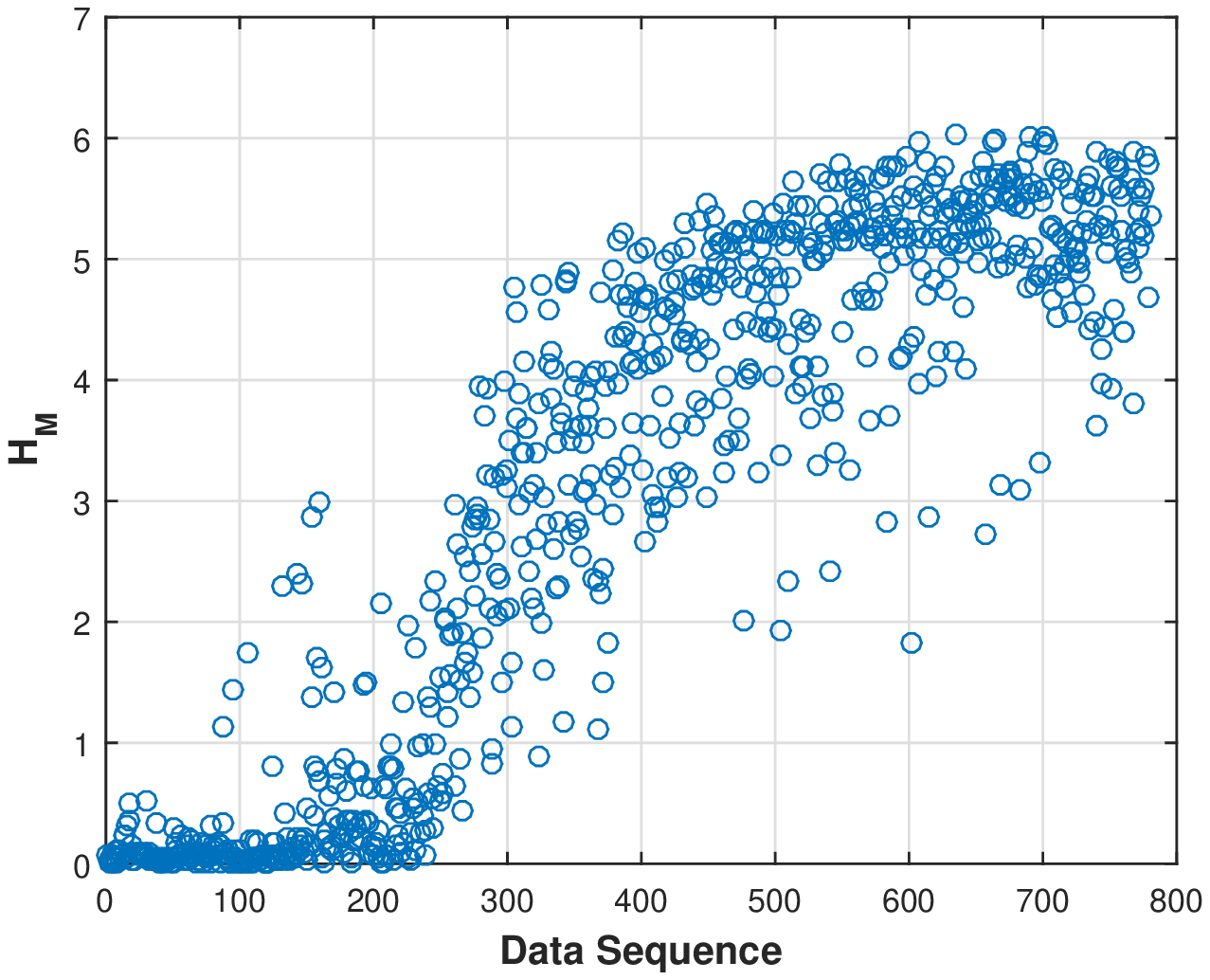}
	\caption{Variation of discrepancy statistics $H_{\mathcal M}$ with increasing pressure fluctuations. This also shows an anomaly around the point $200$ and qualitatively agrees to the behavior of $\Delta_{\mathcal M}$.}
	\label{fig:InfoGain}
\end{figure}

The statistics for the discrepancy measure for the full state models is shown in Figure~\ref{fig:InfoGain}. The plot in Figure~\ref{fig:InfoGain} also agrees qualitatively with the earlier results on $\Delta_{\mathcal M}$. From these plots, we can infer that the Markov statistics for the stable cluster is very similar to the i.i.d. statistics and thus the data is very much independently distributed and conditioning on the inferred states of the Markov models doesn't improve predictability (or information complexity) of the temporal model. Thus, these two measures help infer the changes in the behavior of the data during the combustion process and are useful for anomaly detection.
\begin{figure} %Fig11 
	\centering
	\subfloat[$\Delta_{\mathcal{M}}$ for the full state model for the time-series data with increasing pressure root mean square]{\includegraphics[width=0.75\textwidth]{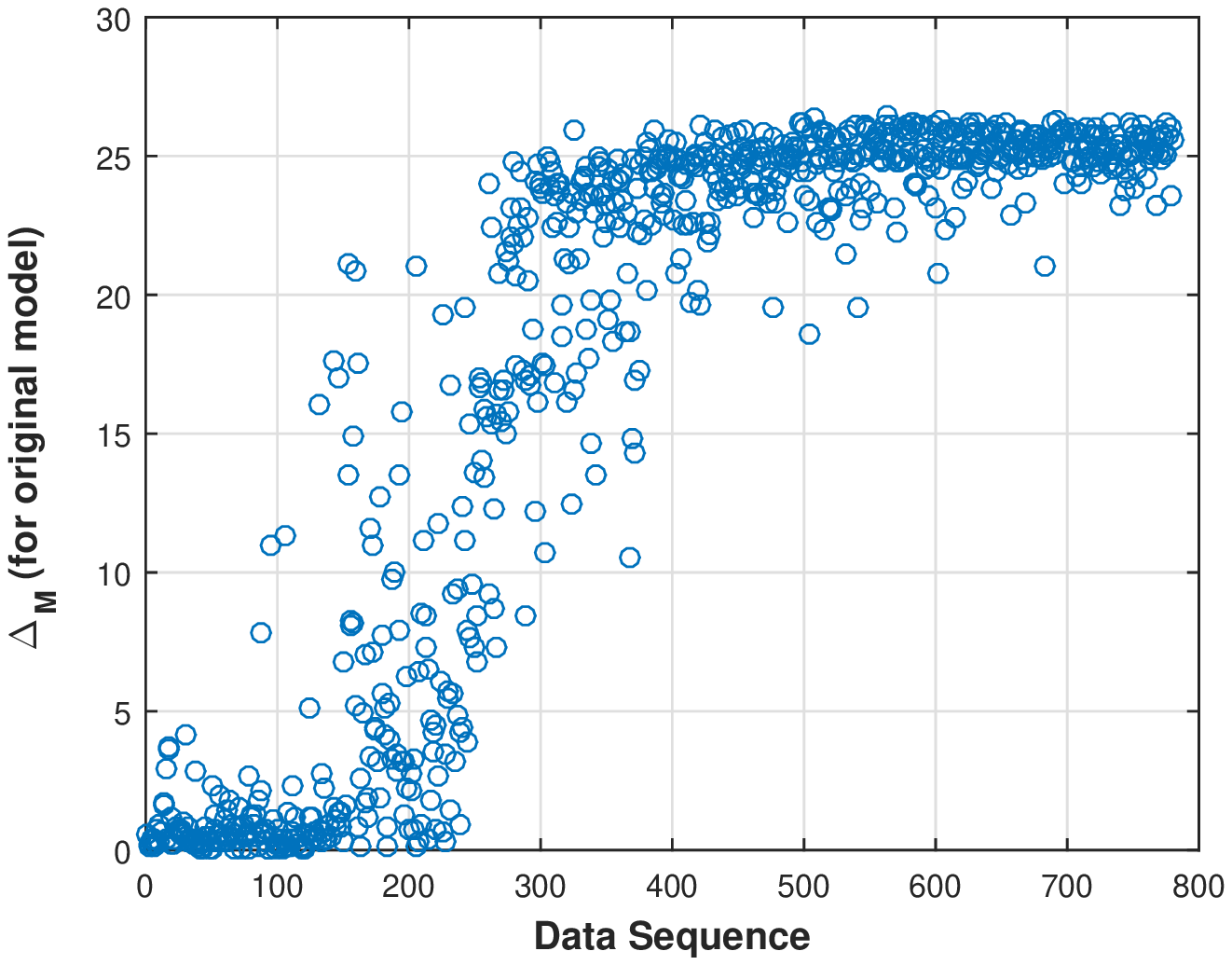}\label{fig:complexorig}}\\
	\subfloat[Typical pressure signals from the three clusters seen in Figure~\ref{fig:complexorig}]{\includegraphics[width=0.75\textwidth]{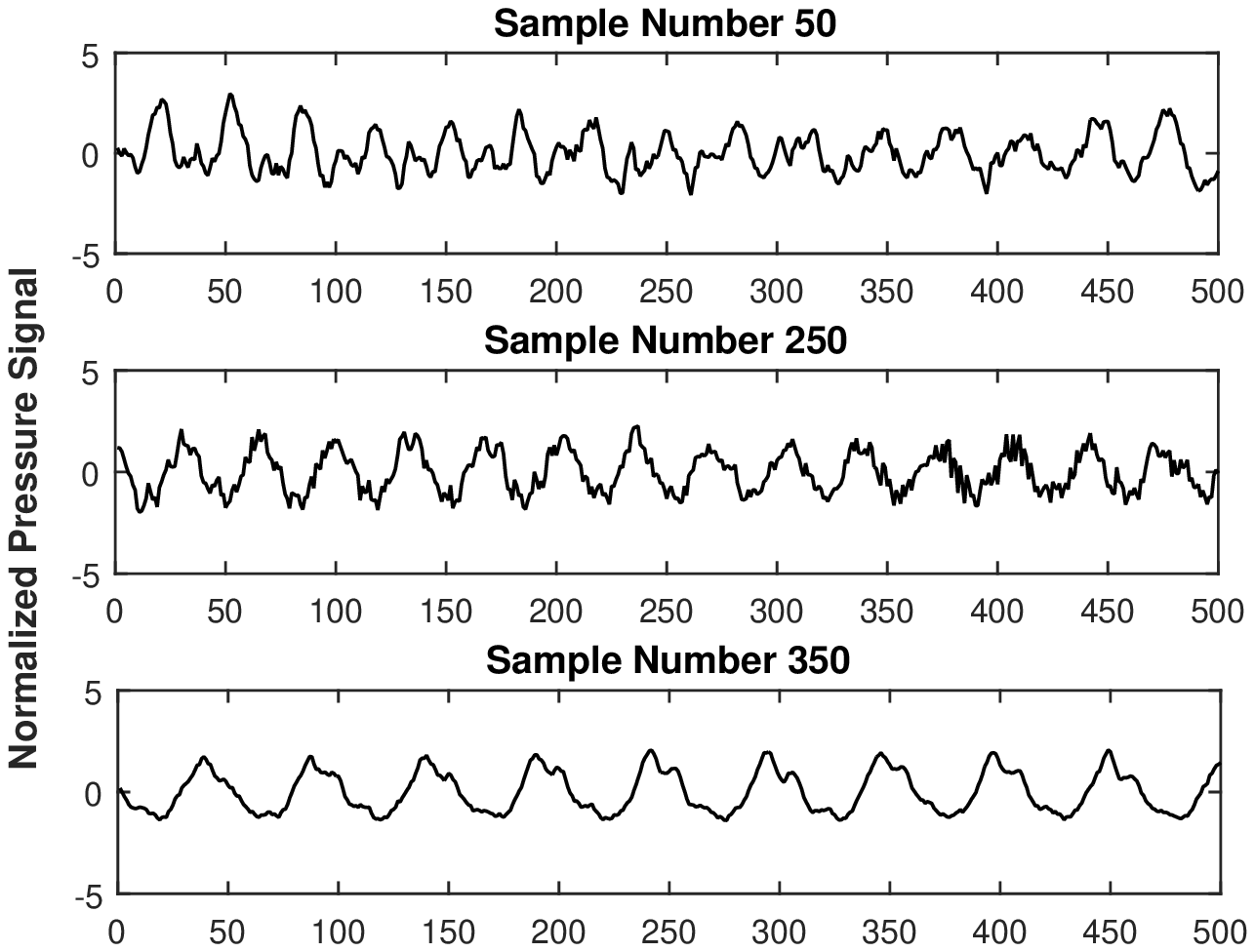}\label{fig:pressure}}
\end{figure}
\begin{figure}\ContinuedFloat %% split over multiple pages for arxiv
  	\centering
	\subfloat[$\Delta_{\mathcal{M}}$ for models with $4$ states for the time-series data with increasing pressure root mean square]{\includegraphics[width=0.75\textwidth]{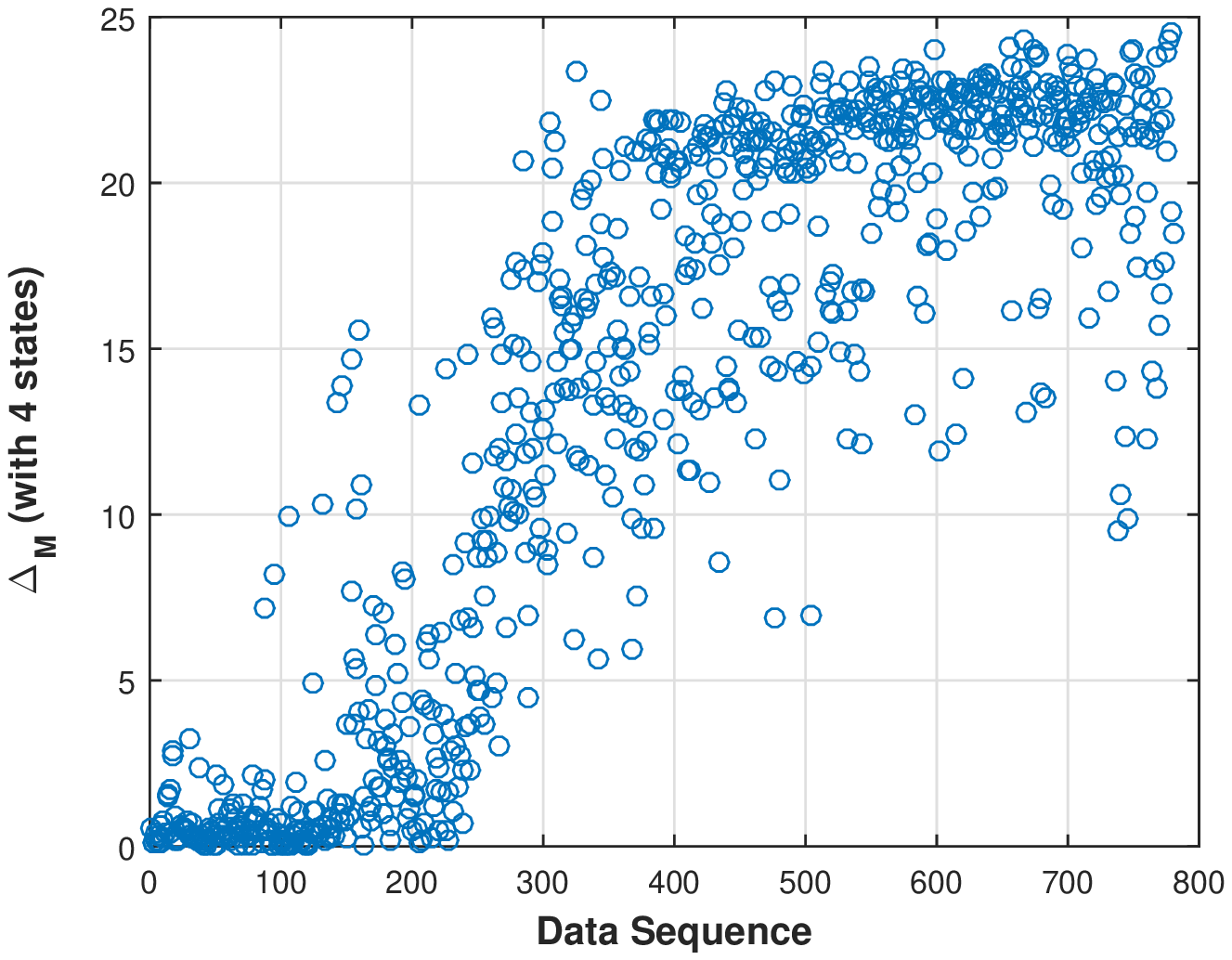}\label{fig:complexfinal}}
%	\subfloat[Histogram for the measure $\mathfrak{d}$ for the reduced model]{\includegraphics[width=0.75\textwidth]{figures/fig9d.eps}\label{fig:complexfinal}}\\
	\caption{Anomalous behavior of data in the combustion process}
	\label{fig:divergence}\vspace{-2pt}
\end{figure}

\begin{figure} %Fig13
	\centering
	\includegraphics[width=0.75\textwidth]{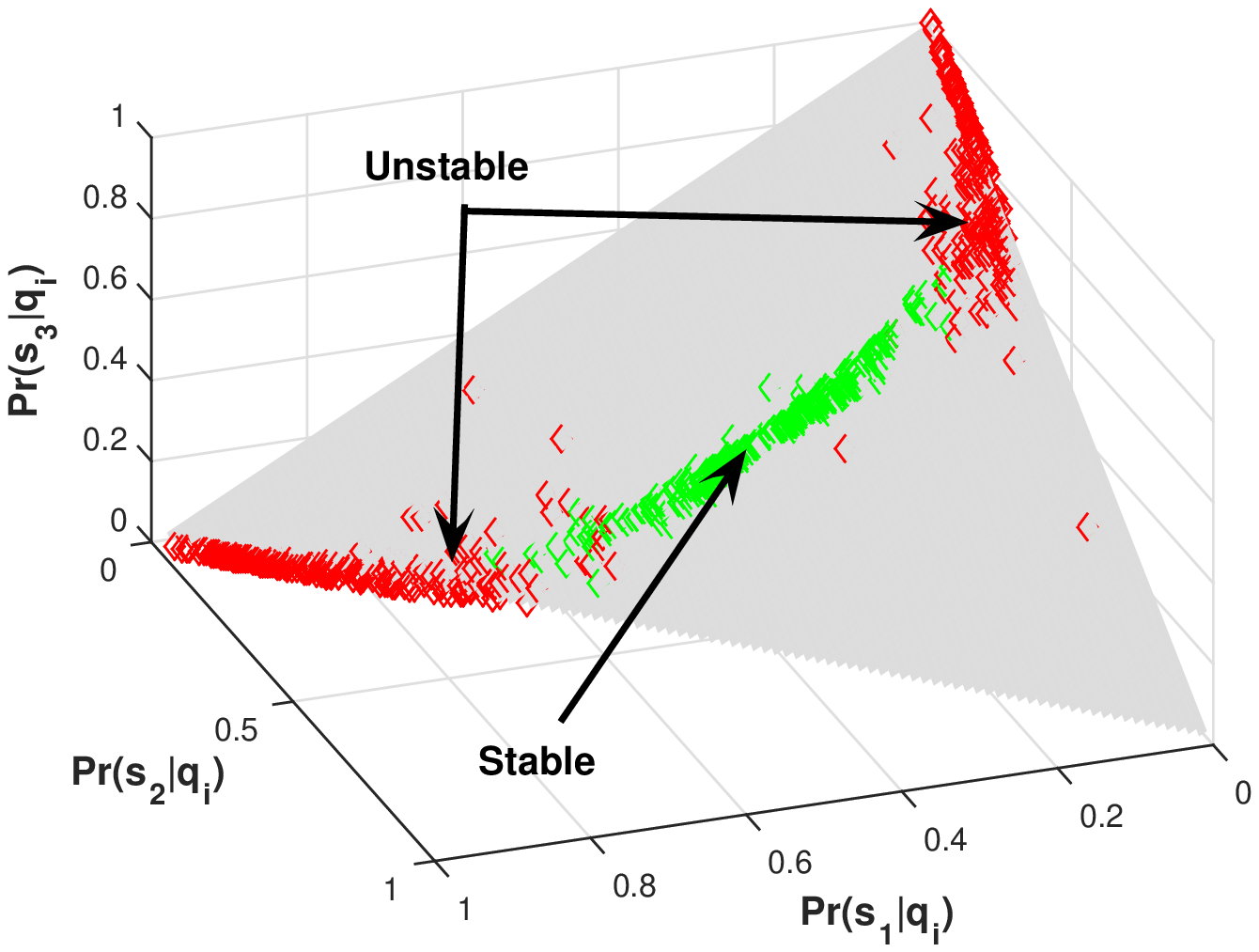}
	\caption{Cluster of stable and unstable phase in information space. Each point is a row of the emission matrix for the reduced Markov model with $2$ states. The plot shows the change in the Markov model as the process moves from stable and unstable. \color{red} Red diamonds \color{black} represent the unstable phase while \color{green} green diamonds \color{black} represent the stable phase.}
	\label{fig:Simplex}
\end{figure}

To see more explicitly the changes in the underlying models, the models during stable and unstable phases are visualized in the information space. To do this, we reduce the state space of the models to just $2$ states and estimate the corresponding emission parameters. As the models have three symbols, the emission matrix has $2$ rows and each row corresponds to the symbol emission probabilities conditioned on the two states. Each of these rows for $100$ cases from stable and $100$ cases from unstable are plotted on a single simplex plane which is shown in Figure~\ref{fig:Simplex}. The Figure shows the clusters of stable and unstable cases in the information space and that the model with even $2$ states are clustered separately. This shows that there is a structured change in the temporal dynamics of the data at the two phases and that the inferred Markov models are able to capture this change. Furthermore, the distinctive features of the models are sufficiently retained even after significant reduction in the state-space of the models.

\subsection{Classification}
These models are then used to train classifiers using support vector machines (SVM) and decision trees (DT)~\cite{B06}. The rationale behind using multiple classifier is to show that the performance of the Markov models is independent of the classification technique (i.e., it works equally well with maximum margin classifiers or decision tree classifiers). The SVM classifier is trained using a radial basis function kernel while the decision tree is trained using the standard euclidean distance. The classifiers are trained with $100$ data points from each class and are tested on the remaining data (around $80$ and $380$ for stable and unstable respectively). The tests are repeated for $100$ different train and test data sets from the total data. The results of classification accuracy are listed in Table~\ref{table:classification}. The SVM classifier is able to achieve around $1.67\%$ error using models with $2$ states while the decision tree classifier is able to achieve around $4.70\%$ error using models with $4$ states. This provides another way of selecting the final model for state merging in a supervised learning setting. It is noted that the original models contain $9$ states for stable and $27$ for unstable class.
\begin{table}
  \caption{Performance of classifiers with different number of states. Mean Error= Lower is better. }\label{table:classification}
  \centering
  \begin{tabular}{cccc}
    \hline
    Number of States & Classifier & Classification Error ($\%$)\\
    \hline
    \multirow{2}{*}{$9$} & SVM & $3.48 \pm 0.74$ \\
    & DT & $9.83\pm 3.24$ \\
    \hline
    \multirow{2}{*}{$8$} & SVM & $3.62 \pm 0.71$ \\
    & DT & $9.38 \pm 3.11$ & \\
    \hline
    \multirow{2}{*}{$7$} & SVM & $2.87 \pm 0.68$ \\
    & DT & $7.70 \pm 2.61$ & \\
    \hline
    \multirow{2}{*}{$6$} & SVM & $2.48 \pm 0.61$ \\
    & DT & $7.00 \pm 2.55$ & \\
    \hline
    \multirow{2}{*}{$5$} & SVM & $2.05 \pm 0.54$ \\
    & DT & $6.10 \pm 2.17$ & \\
    \hline
    \multirow{2}{*}{$4$} & SVM & $1.86 \pm 0.43$ \\
    & DT & $4.72 \pm 2.29$ & \\
    \hline
    \multirow{2}{*}{$3$} & SVM & $1.69 \pm 0.45$ \\
    & DT & $5.56 \pm 1.90$ & \\
    \hline
    \multirow{2}{*}{$2$} & SVM & $1.67 \pm 0.43$ \\
    & DT & $4.83 \pm 1.80$ & \\
    \hline
  \end{tabular}
\end{table}

\section{Summary, Conclusions and Future Work}
In recent times the idea of representation learning has become very popular in the machine learning literature as it allows decoupling of data for model learning  from the end-objectives like classification or clustering. In this paper, we presented a technique for Markov modeling of time-series data using concepts of symbolic dynamics which allows inference of model structure as well as parameters for compact data representation. In the proposed technique we first estimate the memory size of the discretized time-series data. The size of memory is estimated using spectral decomposition properties of the one-step Markov model created from the symbol sequence. Then, a second pass of data is made to infer the model with the right memory and the corresponding symbol emission matrix is estimated. Then the equivalence class of states based on K-L distance between the states are estimated using hierarchical clustering of the corresponding states of the Markov model. The proposed concepts were validated using two different datasets-- combustion instability and bearing. Modeling of combustion instability still remains a puzzle in the combustion community. The Markov modeling technique was used to analyze the problem of combustion instability. The proposed ideas were tested on experimental data from a swirl-stabilized combustor used to study unstable thermo-acoustic phenomenon during combustion process. The proposed approach allows us to infer the complexity of the time-series data based on the inferred Markov model. Two different metrics were proposed for anomaly detection and classification of the stable and unstable classes. The results presented in this paper are encouraging as the inferred models are able to identify the stable and unstable phases independent of any other operating condition.

Simultaneous optimization of discretization and memory estimation for model inference is a topic of future research. While the results obtained with Markov modeling for the combustion instability problem are inspiring, further investigation with transient data is required for better characterization of the process. More thorough comparison of the proposed models with HMM models of similar state-space size is also an important topic of future work.
 
\section*{Acknowledgments}
The authors would like to thank Professor Domenic Santavicca and Mr. Jihang Li of Center for Propulsion, Penn State for kindly providing the experimental data for combustion used in this work.

\bibliographystyle{IEEEtranS}
\bibliography{references.bib}

% Generated by IEEEtranS.bst, version: 1.14 (2015/08/26)
\begin{thebibliography}{10}
\providecommand{\url}[1]{#1}
\csname url@samestyle\endcsname
\providecommand{\newblock}{\relax}
\providecommand{\bibinfo}[2]{#2}
\providecommand{\BIBentrySTDinterwordspacing}{\spaceskip=0pt\relax}
\providecommand{\BIBentryALTinterwordstretchfactor}{4}
\providecommand{\BIBentryALTinterwordspacing}{\spaceskip=\fontdimen2\font plus
\BIBentryALTinterwordstretchfactor\fontdimen3\font minus
  \fontdimen4\font\relax}
\providecommand{\BIBforeignlanguage}[2]{{%
\expandafter\ifx\csname l@#1\endcsname\relax
\typeout{** WARNING: IEEEtranS.bst: No hyphenation pattern has been}%
\typeout{** loaded for the language `#1'. Using the pattern for}%
\typeout{** the default language instead.}%
\else
\language=\csname l@#1\endcsname
\fi
#2}}
\providecommand{\BIBdecl}{\relax}
\BIBdecl

\bibitem{NASAPHM}
\BIBentryALTinterwordspacing
Prognostic data repository: Bearing data set nsf i/ucrc center for intelligent
  maintenance systems, 2010. [Online]. Available:
  \url{http://ti.arc.nasa.gov/tech/dash/pcoe/prognostic-data-repository/}
\BIBentrySTDinterwordspacing

\bibitem{Akaike:1974a}
H.~Akaike, ``A new look at the statistical model identification,'' \emph{IEEE
  Transactions on Automatic Control}, vol.~19, no.~6, pp. 716--723, Dec 1974.

\bibitem{BMH07}
A.~Banaszuk, P.~G. Mehta, and G.~Hagen, ``The role of control in design: From
  fixing problems to the design of dynamics,'' \emph{Control Engineering
  Practice}, vol.~15, no.~10, pp. 1292--1305, 2007.

\bibitem{BMJK06}
A.~Banaszuk, P.~G. Mehta, C.~A. Jacobson, and A.~I. Khibnik, ``Limits of
  achievable performance of controlled combustion processes,'' \emph{Control
  Systems Technology, IEEE Transactions on}, vol.~14, no.~5, pp. 881--895,
  2006.

\bibitem{B06}
C.~M. Bishop, \emph{Pattern recognition and machine learning}.\hskip 1em plus
  0.5em minus 0.4em\relax Springer, 2006.

\bibitem{CDSBM14}
S.~Candel, D.~Durox, T.~Schuller, J.-F. Bourgouin, and J.~P. Moeck, ``Dynamics
  of swirling flames,'' \emph{Annual review of fluid mechanics}, vol.~46, pp.
  147--173, 2014.

\bibitem{CL13}
I.~Chattopadhyay and H.~Lipson, ``Abductive learning of quantized stochastic
  processes with probabilistic finite automata,'' \emph{Philosophical
  Transactions of the Royal Society of London A: Mathematical, Physical and
  Engineering Sciences}, vol. 371, no. 1984, p. 20110543, 2013.

\bibitem{D05}
F.~Darema, ``Dynamic data driven applications systems: New capabilities for
  application simulations and measurements,'' in \emph{5th International
  Conference on Computational Science - ICCS 2005}, Atlanta, GA; United States,
  2005.

\bibitem{GPKK15}
B.~C. Geiger, T.~Petrov, G.~Kubin, and H.~Koeppl, ``Optimal kullback--leibler
  aggregation via information bottleneck,'' \emph{Automatic Control, IEEE
  Transactions on}, vol.~60, no.~4, pp. 1010--1022, 2015.

\bibitem{G90}
R.~M. Gray, \emph{Entropy and information}.\hskip 1em plus 0.5em minus
  0.4em\relax Springer, 1990.

\bibitem{HY09}
Y.~Huang and V.~Yang, ``Dynamics and stability of lean-premixed
  swirl-stabilized combustion,'' \emph{Progress in Energy and Combustion
  Science}, vol.~35, no.~4, pp. 293--364, 2009.

\bibitem{JSR16}
D.~K. Jha, A.~Srivastav, and A.~Ray, ``Temporal learning in video data using
  deep learning and {G}aussian processes,'' in \emph{Workshop on Machine
  Learning for Prognostics and Health Managament at 2016 KDD, San Francisco,
  CA}, 2016.

\bibitem{JSMR15}
D.~K. Jha, A.~Srivastav, K.~Mukherjee, and A.~Ray, ``Depth estimation in
  {M}arkov models of time-series data via spectral analysis,'' in
  \emph{American Control Conference (ACC), 2015}.\hskip 1em plus 0.5em minus
  0.4em\relax IEEE, 2015, pp. 5812--5817.

\bibitem{SAX07}
J.~Lin, E.~Keogh, L.~Wei, and S.~Lonardi, ``Experiencing {SAX}: a novel
  symbolic representation of time series,'' \emph{Data Mining and Knowledge
  Discovery}, vol.~15, no.~2, pp. 107--144, October 2007.

\bibitem{LM95}
D.~Lind and B.~Marcus, \emph{An introduction to symbolic dynamics and
  coding}.\hskip 1em plus 0.5em minus 0.4em\relax Cambridge University Press,
  1995.

\bibitem{M96}
K.~Marton, ``Bounding $\bar{d}$-distance by informational divergence: a method
  to prove measure concentration,'' \emph{The Annals of Probability}, vol.~24,
  no.~2, pp. 857--866, 1996.

\bibitem{MBDSC12}
J.~P. Moeck, J.-F. Bourgouin, D.~Durox, T.~Schuller, and S.~Candel, ``Nonlinear
  interaction between a precessing vortex core and acoustic oscillations in a
  turbulent swirling flame,'' \emph{Combustion and Flame}, vol. 159, no.~8, pp.
  2650--2668, 2012.

\bibitem{MR14}
K.~Mukherjee and A.~Ray, ``State splitting and merging in probabilistic finite
  state automata for signal representation and analysis,'' \emph{Signal
  Processing}, vol. 104, pp. 105--119, 2014.

\bibitem{MS15}
M.~Murugesan and R.~Sujith, ``Combustion noise is scale-free: transition from
  scale-free to order at the onset of thermoacoustic instability,''
  \emph{Journal of Fluid Mechanics}, vol. 772, pp. 225--245, 2015.

\bibitem{NTS14}
V.~Nair, G.~Thampi, and R.~Sujith, ``Intermittency route to thermoacoustic
  instability in turbulent combustors,'' \emph{Journal of Fluid Mechanics},
  vol. 756, pp. 470--487, 2014.

\bibitem{OAL15}
J.~O'Connor, V.~Acharya, and T.~Lieuwen, ``Transverse combustion instabilities:
  Acoustic, fluid mechanic, and flame processes,'' \emph{Progress in Energy and
  Combustion Science}, vol.~49, pp. 1--39, 2015.

\bibitem{QLLY06}
H.~Qiu, J.~Lee, J.~Lin, and G.~Yu, ``Wavelet filter-based weak signature
  detection method and its application on rolling element bearing
  prognostics,'' \emph{Journal of sound and vibration}, vol. 289, no.~4, pp.
  1066--1090, 2006.

\bibitem{RR06}
V.~Rajagopalan and A.~Ray, ``Symbolic time series analysis via wavelet-based
  partitioning,'' \emph{Signal Processing}, vol.~86, no.~11, pp. 3309--3320,
  2006.

\bibitem{R04}
A.~Ray, ``Symbolic dynamic analysis of complex systems for anomaly detection,''
  \emph{Signal Processing}, vol.~84, no.~7, pp. 1115--1130, July 2004.

\bibitem{SCRR16}
S.~Sarkar, S.~R. Chakravarthy, V.~Ramanan, and A.~Ray, ``Dynamic data-driven
  prediction of instability in a swirl-stabilized combustor,''
  \emph{International Journal of Spray and Combustion Dynamics}, p.
  1756827716642091, 2016.

\bibitem{Schwarz:1978a}
G.~Schwarz, ``Estimating the dimension of a model,'' \emph{Ann. Statist.},
  vol.~6, no.~2, pp. 461--464, 03 1978.

\bibitem{SSDC03}
S{\'e}, b.~Ducruix, T.~Schuller, D.~Durox, S{\'e}, and b.~Candel, ``Combustion
  dynamics and instabilities: Elementary coupling and driving mechanisms,''
  \emph{Journal of Propulsion and Power}, vol.~19, no.~5, pp. 722--734, 2003.

\bibitem{SS04}
C.~R. Shalizi and K.~L. Shalizi, ``Blind construction of optimal nonlinear
  recursive predictors for discrete sequences,'' in \emph{Proceedings of the
  20th Conference on Uncertainty in Artificial Intelligence}, ser. UAI '04,
  2004, pp. 504--511.

\bibitem{Srivastav2014}
A.~Srivastav, ``Estimating the size of temporal memory for symbolic analysis of
  time-series data,'' \emph{American Control Conference, Portland, OR, USA},
  pp. 1126--1131, June 2014.

\bibitem{TMZT12}
D.~A. Tobon-Mejia, K.~Medjaher, N.~Zerhouni, and G.~Tripot, ``A data-driven
  failure prognostics method based on mixture of gaussians hidden {M}arkov
  models,'' \emph{IEEE Transactions on reliability}, vol.~61, no.~2, pp.
  491--503, 2012.

\bibitem{VTDCC05}
E.~Vidal, F.~Thollard, C.~De~La~Higuera, F.~Casacuberta, and R.~C. Carrasco,
  ``Probabilistic finite-state machines-part i,'' \emph{Pattern Analysis and
  Machine Intelligence, IEEE Transactions on}, vol.~27, no.~7, pp. 1013--1025,
  2005.

\bibitem{V12}
M.~Vidyasagar, ``A metric between probability distributions on finite sets of
  different cardinalities and applications to order reduction,''
  \emph{Automatic Control, IEEE Transactions on}, vol.~57, no.~10, pp.
  2464--2477, 2012.

\bibitem{VJR16}
N.~Virani, D.~K. Jha, and A.~Ray, ``Sequential hypothesis tests using {M}arkov
  models of time series data,'' in \emph{Workshop on Machine Learning for
  Prognostics and Health Managament at 2016 KDD, San Francisco, CA}, 2016.

\bibitem{XW05}
R.~Xu and D.~Wunsch, ``Survey of clustering algorithms,'' \emph{Neural
  Networks, IEEE Transactions on}, vol.~16, no.~3, pp. 645--678, 2005.

\bibitem{XSB14}
Y.~Xu, S.~M. Salapaka, and C.~L. Beck, ``Aggregation of graph models and
  {M}arkov chains by deterministic annealing,'' \emph{Automatic Control, IEEE
  Transactions on}, vol.~59, no.~10, pp. 2807--2812, 2014.

\end{thebibliography}
\end{document}